\documentclass[lettersize,journal]{IEEEtran}

\usepackage{cite}
\usepackage{hyperref}

\usepackage{color}
\ifCLASSINFOpdf
\usepackage[pdftex]{graphicx}
\usepackage{subfigure}
\else
\fi

\usepackage{amsmath}
\usepackage{amsthm}

\usepackage{times}
\usepackage{soul}
\usepackage{url}
\usepackage{orcidlink}
\usepackage{graphicx}
\usepackage{amsmath}
\usepackage{amsthm}
\usepackage{booktabs}
\usepackage{algorithm}
\usepackage{algorithmic}
\usepackage{multirow}
\usepackage{float}  
\usepackage{amsfonts,amssymb}
\usepackage{makecell}
\usepackage{inconsolata}
\usepackage[T1]{fontenc}
\usepackage[table]{xcolor}
\definecolor{softblue}{RGB}{240, 250, 255}
\begin{document}

\title{Constituency Parsing using LLMs}

\author{Xuefeng Bai\orcidlink{0000-0001-7044-0683}, Jialong Wu\orcidlink{0000-0002-0223-6792}, Yulong Chen\orcidlink{0000-0003-2386-4656}, Zhongqing Wang\orcidlink{0000-0003-0839-9856},\\ Kehai Chen\orcidlink{0000-0002-4346-7618}, Min Zhang\orcidlink{0000-0002-3895-5510}, Yue Zhang\orcidlink{0000-0002-5214-2268},~\IEEEmembership{Senior Member,~IEEE}
    \thanks{
    \textcircled{c}2020 IEEE.  Personal use of this material is permitted.  Permission from IEEE must be obtained for all other uses, in any current or future media, including reprinting/republishing this material for advertising or promotional purposes, creating new collective works, for resale or redistribution to servers or lists, or reuse of any copyrighted component of this work in other works.
    
	Corresponding authors: Yue Zhang; Kehai Chen.

	Xuefeng Bai, Kehai Chen, and Min Zhang are with the School of Computer Science of Technology, Harbin Institute of Technology, Shenzhen 518055, China (email: baixuefeng@hit.edu.cn, chenkehai@hit.edu.cn, zhangmin2021@hit.edu.cn).
	
        Jialong Wu is with the School of Computer Science and Engineering, Southeast University, Nanjing 211189, (email: jialongwu@seu.edu.cn).
        
	Yulong Chen is with the Department of Computer Science and Technology, University of Cambridge, Cambridge, UK, CB3 0FD (email: yc632@cam.ac.uk).
    
        Zhongqing Wang is with the School of Computer Science and Technology, Soochow University, Suzhou 215006, China (email: wangzq@suda.edu.cn).
    
	Yue Zhang is with the School of Engineering, Westlake University, and also with the Institute of Advanced Technology, Westlake Institute for Advanced Study, Hangzhou 310024, China (email: yue.zhang@wias.org.cn).
    }
}



\maketitle

\begin{abstract}
Constituency parsing is a fundamental yet unsolved challenge in natural language processing.
In this paper, we examine the potential of recent large language models (LLMs) to address this challenge.
We reformat constituency parsing as a sequence-to-sequence generation problem and evaluate the performance of a diverse range of LLMs under zero-shot, few-shot, and supervised fine-tuning learning paradigms.
We observe that while LLMs achieve acceptable improvements, 
they still encounter substantial limitations, due to the absence of mechanisms to guarantee the validity and faithfulness of the generated constituent trees.
Motivated by this observation, we propose two strategies to guide LLMs to generate more accurate constituent trees by learning from erroneous samples and 
refining outputs in a multi-agent collaboration way, 
respectively.
The experimental results demonstrate that our methods effectively reduce the occurrence of invalid and unfaithful trees, thereby enhancing overall parsing performance and achieving promising results across different learning paradigms. 
\end{abstract}


\section{Introduction}
\begin{figure}
    \centering
    \includegraphics[width=0.35\textwidth]{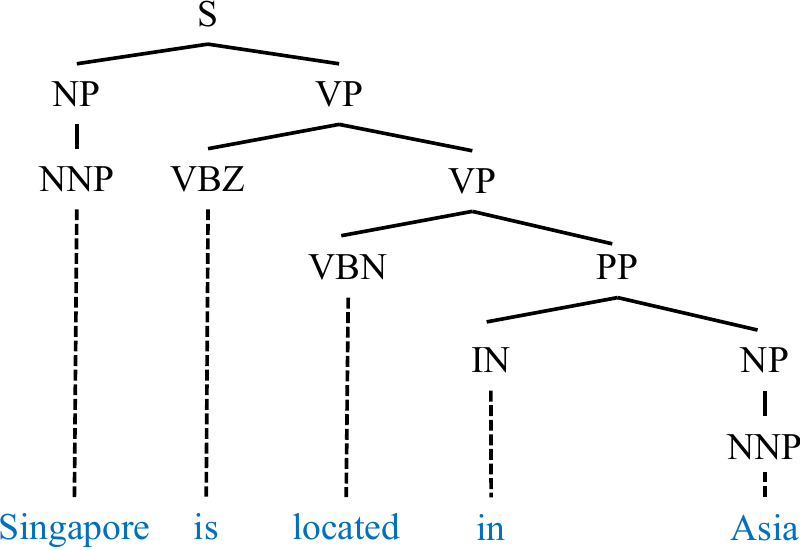}
    \caption{The constituency tree for sentence ``\textcolor[HTML]{0070c0}{Singapore is located in Asia.}''.}
    \label{fig:example}
\end{figure}

\IEEEPARstart{C}{onstituency} parsing is a fundamental natural language processing (NLP) task which aims to predict the syntactic structure of a given sentence according to a phrase structure grammar~\cite{10.5555/972470.972475,collins-1997-three,charniak-2000-maximum, petrov-klein-2007-improved}.
As shown in Figure~\ref{fig:example}, given the sentence ``\textit{Singapore is located in Asia.}'', a constituent parser produces its phrase structure tree automatically.
This task is a fundamental problem in computational linguistics studies. 
It also serves as a foundation for improving various natural language processing tasks over the past decades, facilitating numerous downstream applications~\cite{wang-etal-2018-tree,jiang-diesner-2019-constituency,xu-durrett-2019-neural}.
Despite its importance in linguistics and NLP, constituency parsing remains a challenge, with the supervised performance being 96\%~\cite{kitaev-etal-2019-multilingual}, whereas unsupervised approaches lag significantly, with performance falling below 65\%~\cite{gu-etal-2022-phrase}.

Recently, large language models (LLMs) have achieved remarkable performance on many NLP tasks, such as text classification~\cite{chen-etal-2022-adaprompt, SunL0WGZ023}, commonsense reasoning~\cite{zhang2022task,toroghi-etal-2024-verifiable}, text summarization~\cite{goyal2022news, chen2022unisumm,10.1162/tacl_a_00632}, machine translation~\cite{jiao2023chatgpt} and dialogue systems~\cite{qin2023chatgpt}, using prompting technologies.
The primary idea is to convert NLP tasks into a text-to-text problem, enabling knowledge transfer from pre-training tasks to downstream tasks.
However, due to the fact that the output of constituency parsing is represented in a tree structure, which is rarely found in textual data where LLMs are trained, there has been limited research utilizing LLMs for constituency parsing.
It remains an open question whether LLMs with rich linguistic knowledge can generalize well on this task.

To fill this gap, we take the first step to study the potential of LLMs on constituency parsing.
While the current state-of-the-art constituency parsers are graph-based (i.e., predicting constituent spans on pretrained encoders), LLMs are sequence-based in nature.
To enable LLMs to process constituency trees, we first transform constituency trees into sequences of symbols using three linearization strategies.
Subsequently, we empirically evaluate LLMs and compare them with the state-of-the-art constituency parsers~\cite{kitaev-etal-2019-multilingual} across one in-domain dataset and five out-of-domain datasets under zero-shot, few-shot, and supervised fine-tuning settings. 
Experimental results and further analysis demonstrate that:
1) LLMs significantly outperform previous \textit{sequence-based} constituency parsing methods across different settings, obtaining promising results on unsupervised constituency parsing;
2) LLMs underperform \textit{graph-based} methods under supervised fine-tuning settings;
3) While achieving promising performance, LLM-based parsers frequently generate invalid and unfaithful constituent trees, which significantly limit the overall performance.

Based on these, we propose two strategies to mitigate the above limitation and improve the performance of LLMs on constituency parsing. The first strategy, termed \textit{learning from erroneous samples} (LES), involves integrating erroneous samples along with their corresponding error annotations into prompts during inference. 
LES guides LLMs to gain information on how to avoid generating invalid and unfaithful constituent trees. 
The second strategy, termed \textit{parsing with multi-agent collaboration} (PMC), utilizes two additional agents to assess and provide feedback on the validity and faithfulness of the output. 
These feedbacks are subsequently integrated as input context for the LLM-based parsing agent, facilitating the refinement of the generated constituency tree.
We implement the proposed strategies on both closed-source and open-source LLMs and examine their effectiveness on six datasets under both zero-shot/few-shot and supervised fine-tuning settings.
Experimental results demonstrate that the proposed strategies can effectively reduce the generation of invalid and unfaithful constituent trees. 
Consequently, our methods substantially enhance the performance of current LLMs under all three learning settings.
Remarkably, our method yields an F1 score of 76.87 under zero-shot settings, setting a new benchmark for end-to-end unsupervised constituency parsing.
{To summarize, this paper makes the following contributions:
\begin{itemize}
    \item We systematically study the effectiveness of current LLMs on constituency parsing, highlighting the strengths and weaknesses of LLM-based parsers.
    \item Based on our observation, we introduce two strategies to mitigate the generation of invalid and unfaithful trees by learning from erroneous samples and refining outputs in a multi-agent collaboration way.
    \item Experimental results show that the proposed method remarkably reduces the occurrence of invalid and unfaithful trees and enhances overall parsing performance.\footnote{Our code will be publicly available at GitHub.}
\end{itemize}
}

\section{Related Work}
\noindent\textbf{Constituency Parsing}.
Current approaches to constituency parsing are predominantly divided into three categories: graph-based~\cite{collins1999statistical, durrett2015neural}, transition-based~\cite{yamada-matsumoto-2003-statistical,sagae-lavie-2005-classifier,nivre-mcdonald-2008-integrating,zhang-clark-2009-transition}, and sequence-based methods~\cite{vinyals2015grammar,kamigaito-etal-2017-supervised,DBLP:conf/aaai/LiuZS18}.
Graph-based methods assign scores to all spans within a sentence, then employ the CKY dynamic programming algorithm to identify the optimal parse tree~\cite{kitaev-klein-2018-constituency,kitaev-etal-2019-multilingual,zhang-etal-2020-fast}. 
Transition-based methods function by sequentially processing input sentences and progressively building the resultant constituency trees through a succession of locally predicted transition actions~\cite{dyer-etal-2016-recurrent,cross-huang-2016-span,TACL1199}. 
Sequence-based methods solve constituency parsing as a sequence-to-sequence generation problem~\cite{fernandez-gonzalez-gomez-rodriguez-2020-enriched,nguyen-etal-2021-conditional,yang-tu-2022-bottom}. 
This work belongs to the category of sequence-based methods.
Different from existing methods, our research employs generative LLMs to address the task of constituency parsing. 
A recent related work is~\cite{tian-etal-2024-large}, which observes that LLMs are weak in predicting deeper trees and proposes a three-step task composition framework to improve LLMs' capacity for predicting deep tree structures.
{In contrast, our work conducts a systematic examination of LLMs' capabilities and constraints in sequence-based constituency parsing, with a particular focus on structural and semantic fidelity in parse tree generation. In addition, we introduce two novel methodological solutions aimed at improving the validity and faithfulness of LLM-generated parse trees.}

\noindent\textbf{Large Language Models and Structured Prediction}.
Large Language Models (LLMs)~\cite{BrownMRSKDNSSAA20,chowdhery2022palm,touvron2023llama} have substantially influenced the field of natural language processing.
As the pioneering work, Radford \textit{et al}~\cite{radford2019language} and Brown et al~\cite{BrownMRSKDNSSAA20} demonstrate the capability of language models to solve a task with minimal task supervision.
The following work shows that LLMs are adept at leveraging textual instructions to perform various tasks~\cite{sanh2022multitask,webson-pavlick-2022-prompt,liang2022holistic,zhang2022opt}.
LLMs with Reinforcement Learning from Human Feedback (RLHF) can further generate a response that is well aligned with human values~\cite{ouyang2022training,lambert2022illustrating,bai2022training,gao2023scaling}.
LLMs now showcase remarkable performance on a wide array of tasks, including commonsense reasoning~\cite{zhang2022task,toroghi-etal-2024-verifiable}, text summarization~\cite{goyal2022news, chen2022unisumm,10.1162/tacl_a_00632}, and massive multitask language understanding~\cite{hendryckstest2021}.
Inspired by these, recent research investigates LLMs on structure prediction tasks such as information extraction~\cite{XuCPZXZWZWC24,LiZZRLSGLLH0LLY24}, event extraction~\cite{simon-etal-2024-generative,meng-etal-2024-cean}, knowledge graph construction~\cite{ZhuWCQOYDCZ24}, and semantic parsing~\cite{SchneiderKJSM24}.
However, there has been limited research utilizing LLMs for constituency parsing, which presents a unique challenge due to the difficulty in predicting its nested, structured output.
We fill this gap by conducting a comprehensive study on the potential and limitations of LLMs for this task, and by introducing two strategies to enhance their performance.

\section{Constituency Parsing with LLMs}
{
\begin{table}[t]
\centering
\small
{
\caption{List of abbreviations.}
}
\begin{tabular}{ll}
\toprule
\textbf{Abbreviation}                  & \textbf{Definition}  \\
\midrule
LLMs                  & Large Language Models  \\
NLP                   & Natural Language Processing  \\
CKY                     & Cocke-Kasami-Younger algorithm \\
Seq2Seq                     & Sequence-to-Sequence \\
RLHF                     & Reinforcement Learning from Human Feedback \\
PTB                     & Penn Treebank \\
MCTB                     & Multi-domain Constituent Treebank \\
GPT                    & Generative Pre-trained Transformer \\
OPT                     & Open Pre-trained Transformer \\
LSTM                    & Long Short-Term Memory \\
GCN                     &Graph Convolutional Network \\
LP                    &  Labeled Precision \\
LR                    & Labeled Recall \\
LES                    & Learning from Erroneous Samples  \\
PMC                    & Parsing with Multi-agent Collaboration \\
\bottomrule
\end{tabular}
\label{tab:abbreviations}
\vspace{-1.0em}
\end{table}
}

{In this paper, we adopt several abbreviations for technical terms. 
To ensure clarity and consistency, we provide a comprehensive list of abbreviations in Table~\ref{tab:abbreviations}. 

To investigate the applicability of LLMs in constituency parsing, we frame this task as a sequence-to-sequence problem such that constituency trees can be generated autoregressively.
We first adapt three linearization strategies to transform constituency trees into sequences of symbols (\S\ref{sec:linearization}).
To study the zero-shot/few-shot learning abilities of LLMs, we design prompts to guide them to generate outputs without parameter update~(\S\ref{sec:prompting}).
In addition, we perform fine-tuning on the full training dataset to exploit the potential of open-source LLMs on constituency parsing~(\S\ref{sec:tuning}).

\begin{figure}
    \centering
    \includegraphics[width=0.42\textwidth]{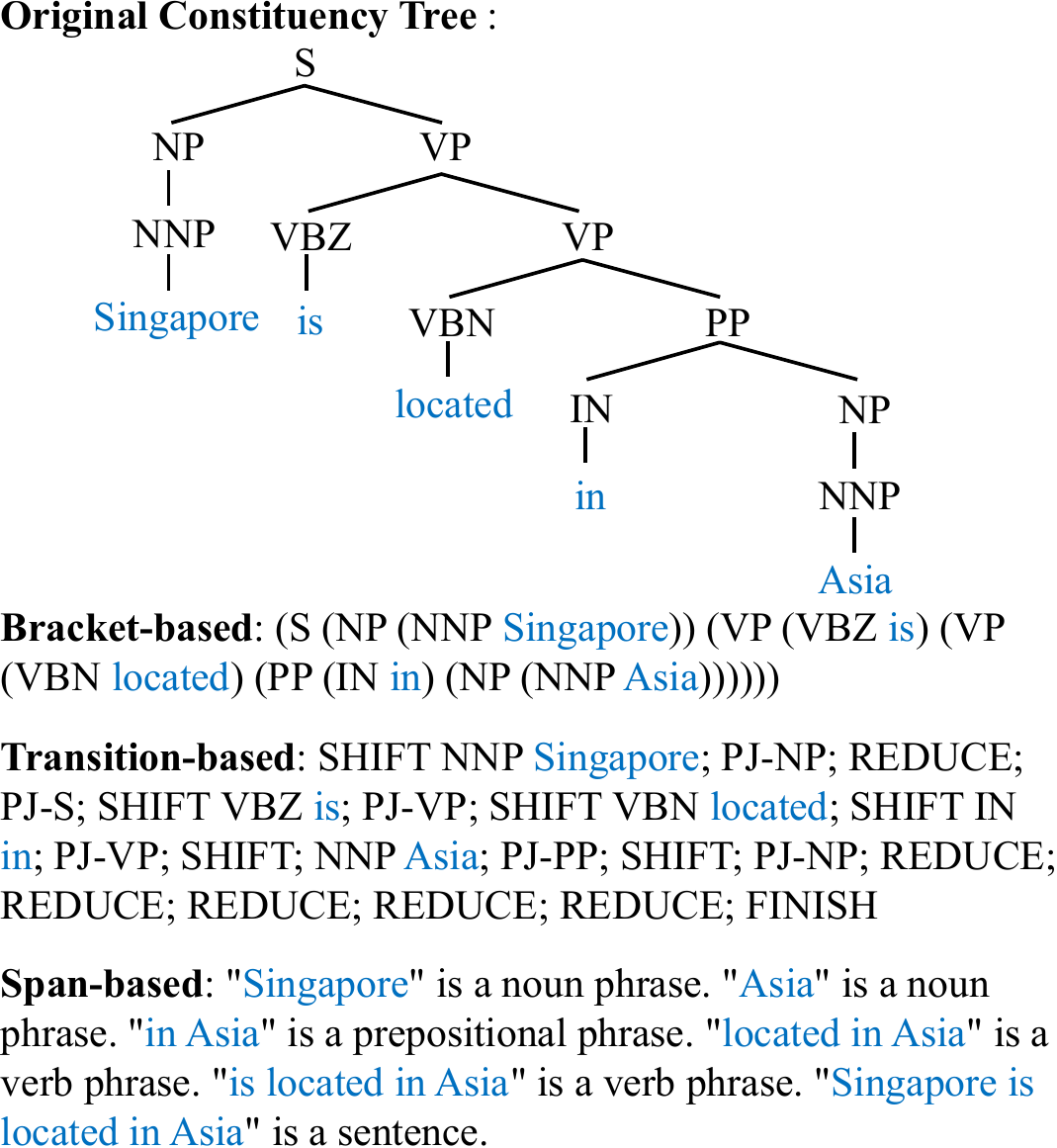}
    \caption{The constituency tree for ``\textcolor[HTML]{0070c0}{Singapore is located in Asia.}'' with three different linearization.}
    \label{fig:linearization}
\end{figure}

\subsection{Tree Linearization}

\label{sec:linearization}
We use three tree-isomorphic\footnote{Tree-isomorphic means that it is possible to decode the original tree from the linearized result without losing adjacency information.} linearization strategies to serialize a constituent tree into a sequence so that LLMs can learn constituency parsing by predicting the linearized tree. 
Figure~\ref{fig:linearization} shows the linearized trees of the constituent tree for a sentence ``{\sl Singapore is located in Asia}'' using three different linearization strategies, which we detail below:

\noindent\textbf{Bracket-based Linearization}. We follow  Vinyals et al ~\cite{vinyals2015grammar} and linearize a constituent tree as a bracket sequence from top to bottom. 
The bracket-based linearization uses parentheses to mark visit depth and restore tree structure.
The bracket-based sequence is different from natural language but similar to codes which account for a certain proportion of the pre-training corpus of LLMs.

\noindent\textbf{Transition-based Linearization}. We follow Liu and Zhang~\cite{TACL1199} and convert a constituent tree into a sequence of transition actions. 
The transition sequence assumes there is a stack and a buffer for representing a state, where the stack holds partially built constituents and the buffer contains the next incoming words.
The parse tree can be constructed by executing transition actions, which incrementally consume words from the buffer and build output structures on the stack.
The transition-based sequence contains a set of special tokens (i.e. transition actions) which are new to LLMs and differ from the natural language distribution.

\noindent\textbf{Span-based Linearization}.
We consider a method, transforming a tree into a sequence of short sentences.
Specifically, we extract all phrases as well as their labels from the tree and convert them into short sentences by filling the template ``\texttt{A is a B}'', where ``\texttt{A}'' denotes a phrase and ``\texttt{B}'' is its corresponding label.
Compared with linearized trees using the above two methods, Span-based linearized tree is most similar to natural language but is much longer due to word/phrase repetitions.
\begin{figure*}[!t]
    \centering
    \includegraphics[width=0.8\textwidth]{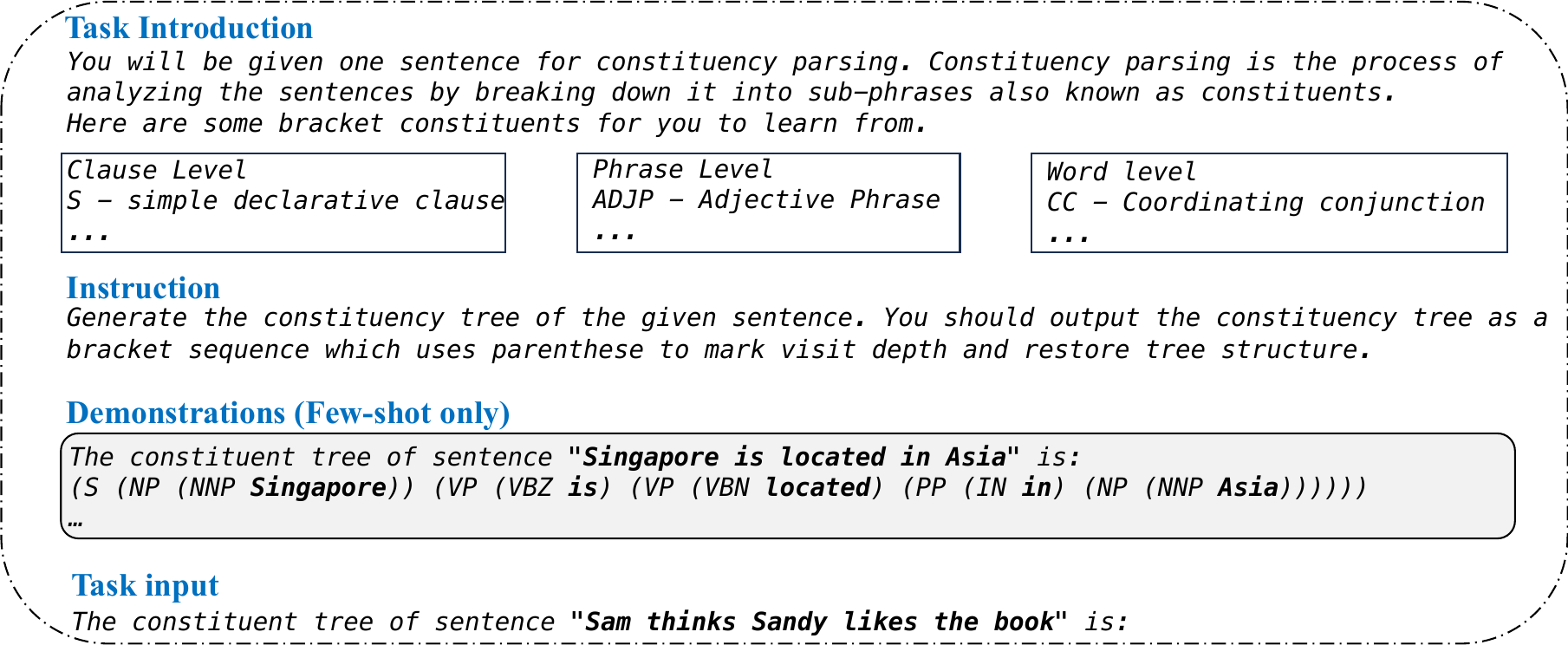}
    \caption{Illustration of prompts for zero/few-shot learning.}
    \label{fig:prompt}
    \vspace{-1.0em}
\end{figure*}

\subsection{Prompting LLMs for Constituency Parsing}
\label{sec:prompting}
To study the capacity of LLMs for zero-shot or few-shot constituency parsing, we follow an instruction-prompt scheme to teach LLMs to generate linearized constituency trees~\cite{ouyang2022training}.
Figure~\ref{fig:prompt} shows the detailed structure of our prompt.
Under the zero-shot setting, our prompt contains three parts: \texttt{Task Introduction}, \texttt{Instruction}, and \texttt{Task Input}.
\texttt{Task Introduction} explains the constituency parsing task and introduces constituent tags, ranging from clause level and phrase level to word level.
\texttt{Instruction} describes the task that needs to be solved and specifies the required output format.
\texttt{Task Input} refers to the input sentence to parse.
Under the few-shot setting, a small number of \texttt{Training Instances} are additionally provided as a context to instruct the LLMs to predict the constituency tree of the input sentence.


\subsection{Tuning LLMs for Constituency Parsing}
\label{sec:tuning}
To explore the full potential of LLMs on constituency parsing, we fine-tune LLMs on the full training dataset.
Specifically, we follow the training paradigm of standard generative language modeling to fine-tune LLMs.
We concatenate \texttt{Instruction}\footnote{We use a short task instruction for fine-tuning, e.g. \textit{Generate the constituency tree of the given sentence.}}, \texttt{Task Input} as well as the linearized constituency tree as a sequence, and train the model to generate the whole sequence incrementally. 

Formally, denoting the concatenated sequence as $Z$, we fine-tune the model by minimizing  the following training objective:
\begin{equation}
	\ell(\theta) = - \sum_{i=1}^{N} \text{log}P_{\theta}(Z_i|Z_{<i}),
\end{equation}
where $Z_i$ is the $i$-th token of output sequence, $Z_{<i}$ is the generated sequence before time step $i$, $\theta$ are model parameters, and $N$ is the length of the output sequence.
Apart from full-parameter fine-tuning, we also explore parameter-efficient learning using low-rank adaptation (LoRA)~\cite{HuSWALWWC22}.

\section{Experiments}
\subsection{Datasets}
\noindent\textbf{In-domain}. 
We use the Penn Treebank (PTB; \cite{10.5555/972470.972475}) dataset for in-domain evaluation, comprising 40k instances from the \textit{news} domain. 
Following standard splits, sections 02–21 are used for training, section 22 for development, and section 23 for testing.


\begin{table}[!t]
\footnotesize
\centering
\caption{Dataset statistics.}
\begin{tabular}{lcccccc}
   \toprule
   \textbf{Dataset} & \multicolumn{1}{c}{\textbf{PTB}} & \multicolumn{5}{c}{\textbf{MCTB}}  \\ 
    \cmidrule(r){1-1} \cmidrule(r){2-2} \cmidrule(r){3-7} 
   {Domain} & News & Dialogue & Forum & Law & Liter. & Review \\
   \midrule
     Train & 39,832 & - & - & - & - & - \\
     Dev & 1,700 & - & - & - & - & - \\
     Test & 2,426 & 1,000 & 1,000 & 1,000 & 1,000 & 1,000 \\
   \bottomrule
\end{tabular}
\label{tab:statistic}
\end{table}

\noindent\textbf{Out-of-domain}.
To investigate the capability of large-scale language models in cross-domain learning scenarios, we conduct an evaluation on the MCTB dataset~\cite{yang-etal-2022-challenges}, which covers five domains: \textit{dialogue}, \textit{forum}, \textit{law}, \textit{literature}, and \textit{review}. 

Table~\ref{tab:statistic} summarizes the statistics of the above datasets.

\subsection{Baselines and basic LLMs}
\noindent The following non-LLM baselines are used for comparison:
\begin{itemize}
    \setlength{\itemsep}{0pt}
    \setlength{\parsep}{0pt}
    \setlength{\parskip}{0pt}
    \item SEPar~\cite{kitaev-klein-2018-constituency} is a chart-based model consisting of a self-attentive encoder and a chart-based decoder.
    \item SAPar~\cite{tian-etal-2020-improving} augments SEPar with the span-based attention mechanism to better learn n-gram features.
    \item TGCN~\cite{yang2020strongly} is a transition-based parser which employs a Graph Convolutional Network (GCN) to encode the partial tree and generate actions. 
    \item LSTM~\cite{vinyals2015grammar} is a sequence-based parser which uses a model with 3 LSTM layers and uses the attention mechanism for better generation.
    \item Vanilla Transformer~\cite{vaswani2017attention} has been employed in parsing, the encoder of which takes the sentence as input while the decoder generates the linearized constituency tree.
    \item GPT-2~\cite{radford2019language} parser is fine-tuned by taking sentences as input and linearized constituency trees as output. We use the large version of GPT-2 for experiments.
    \item PUCP~\cite{cao-etal-2020-unsupervised} is an unsupervised parser that regularizes the parser with phrases from an unsupervised phrase tagger.
    \item UPCT~\cite{gu-etal-2022-phrase} is an unsupervised parser that uses a neural acceptability model to make grammaticality decisions.
\end{itemize}
We experiment with the following LLMs:

\begin{itemize}
    \setlength{\itemsep}{0pt}
    \setlength{\parsep}{0pt}
    \setlength{\parskip}{0pt}
    \item ChatGPT is a conversational version of GPT-3.5 model~\cite{ouyang2022training}. We use the \texttt{turbo} API from OpenAI.\footnote{https://platform.openai.com/docs/models/gpt-3-5}
    \item GPT-4~\cite{openai2023gpt} is the most advanced GPT model from OpenAI. We use the \texttt{gpt-4-turbo} API.\footnote{https://platform.openai.com/docs/models/gpt-4}
    \item OPT~\cite{zhang2022opt} is an LLM pre-trained on 180B tokens.
    \item LLaMA~\cite{touvron2023llama} is an LLM pre-trained on 1.4T\textasciitilde 15T tokens. We use LLaMA-7b, LLaMA3.1-8b-instruct (LLaMA3.1-8b-IT for short), and LLaMA3.3-70b-instruct (LLaMA3.1-70b-IT for short) in experiments.
    \item Vicuna~\cite{vicuna2023} is trained by fine-tuning a LLaMA model on 70K user-shared conversations.
    \item Qwen~\cite{vicuna2023} is an LLM pre-trained on about 18T tokens. We adopt Qwen2.5-7B-instruct (Qwen2.5-8b-IT for short), and Qwen2.5-72B-instruct (Qwen2.5-72b-IT for short) in our experiments.
\end{itemize}

\subsection{Settings}
\noindent\textbf{Pre-Processing}.
Our preliminary experiments show that punctuations often lead to invalid constituency trees.
To reduce such errors, we replace all punctuation marks with special symbols.
More details can be found in our released code.


\noindent\textbf{Evaluation Metric}.
Following previous work~\cite{kitaev-klein-2018-constituency,kitaev-etal-2019-multilingual,zhang-etal-2020-fast,yang-etal-2022-challenges,cui-etal-2022-investigating}, 
we evaluate the model performance using the \texttt{evalb} tool,\footnote{https://nlp.cs.nyu.edu/evalb/} which measures constituent-level labeled precision (LP), recall (LR), and F1 scores (F1). 
It should be noted that we handle invalid constituents differently from the standard \texttt{evalb} script - rather than ignoring them, we assign a score of 0.

\noindent\textbf{Hyper-parameters}. 
We set the decoding temperature as 0 to increase the model’s determinism.
For fine-tuning, we use a learning rate of 3e-5 for 7B and 8B models, and use 2e-5 for 70B and 72B models.
We set the maximum length to 2048.
Different from full parameter fine-tuning, we use a learning rate of 1e-4 for LoRA fine-tuning.
We evaluate the model's performance in the validation dataset after each epoch and choose the model with the highest F1 score as the best model for testing.
Our model is implemented based on \textit{pytorch}\footnote{https://github.com/pytorch/pytorch} and \textit{huggingface transformers}\footnote{\url{https://github.com/huggingface/transformers}} libirary.

\subsection{Results}

\subsubsection{\textbf{Development Experiment Results}}


\begin{figure*}[!t]
	\centering 
	\subfigure[Impact of linearization strategy]{\includegraphics[width=0.3\hsize]{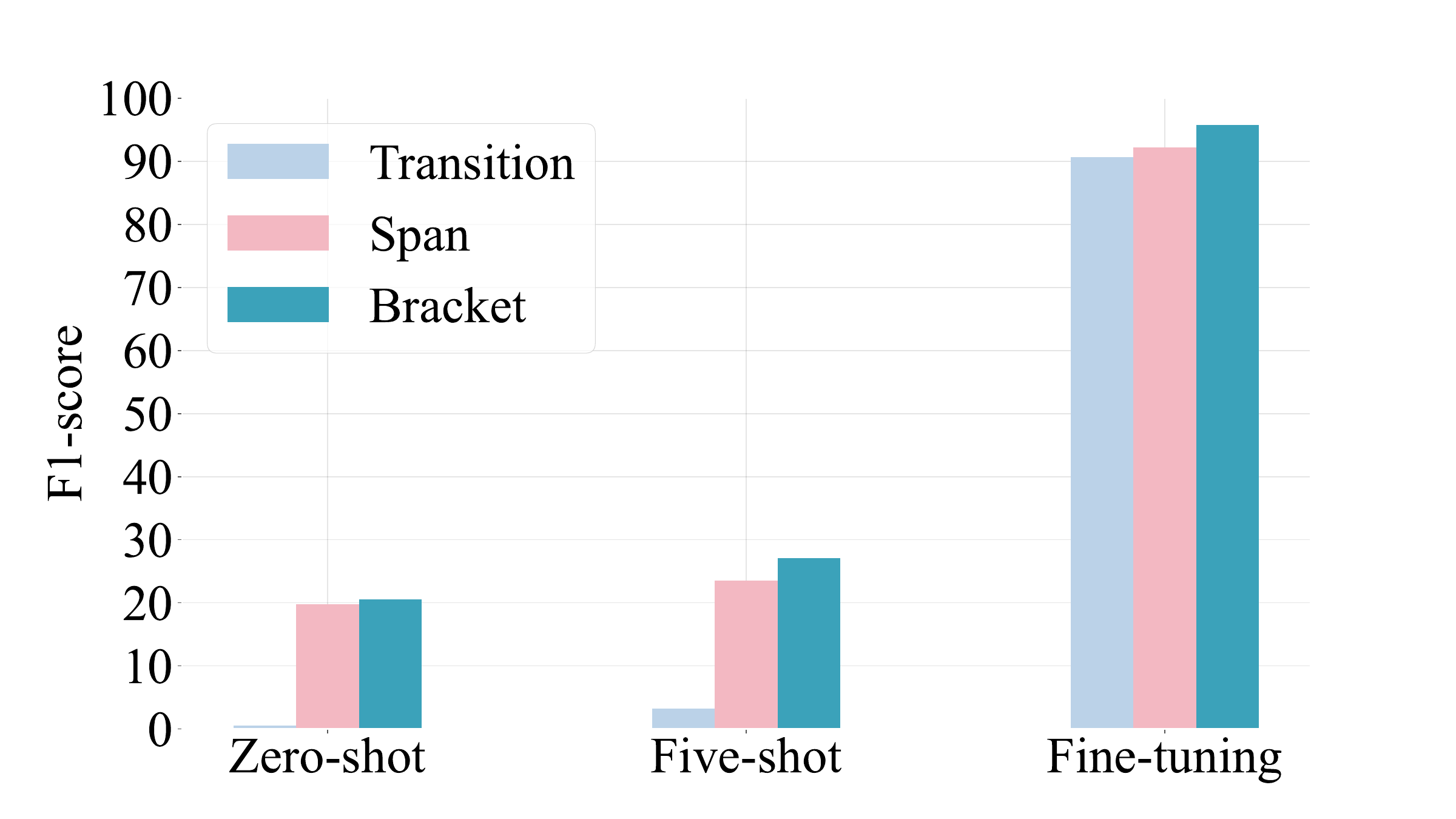}\label{fig:serilization}}
	\subfigure[Impact of decoding strategy]{\includegraphics[width=0.3\hsize]{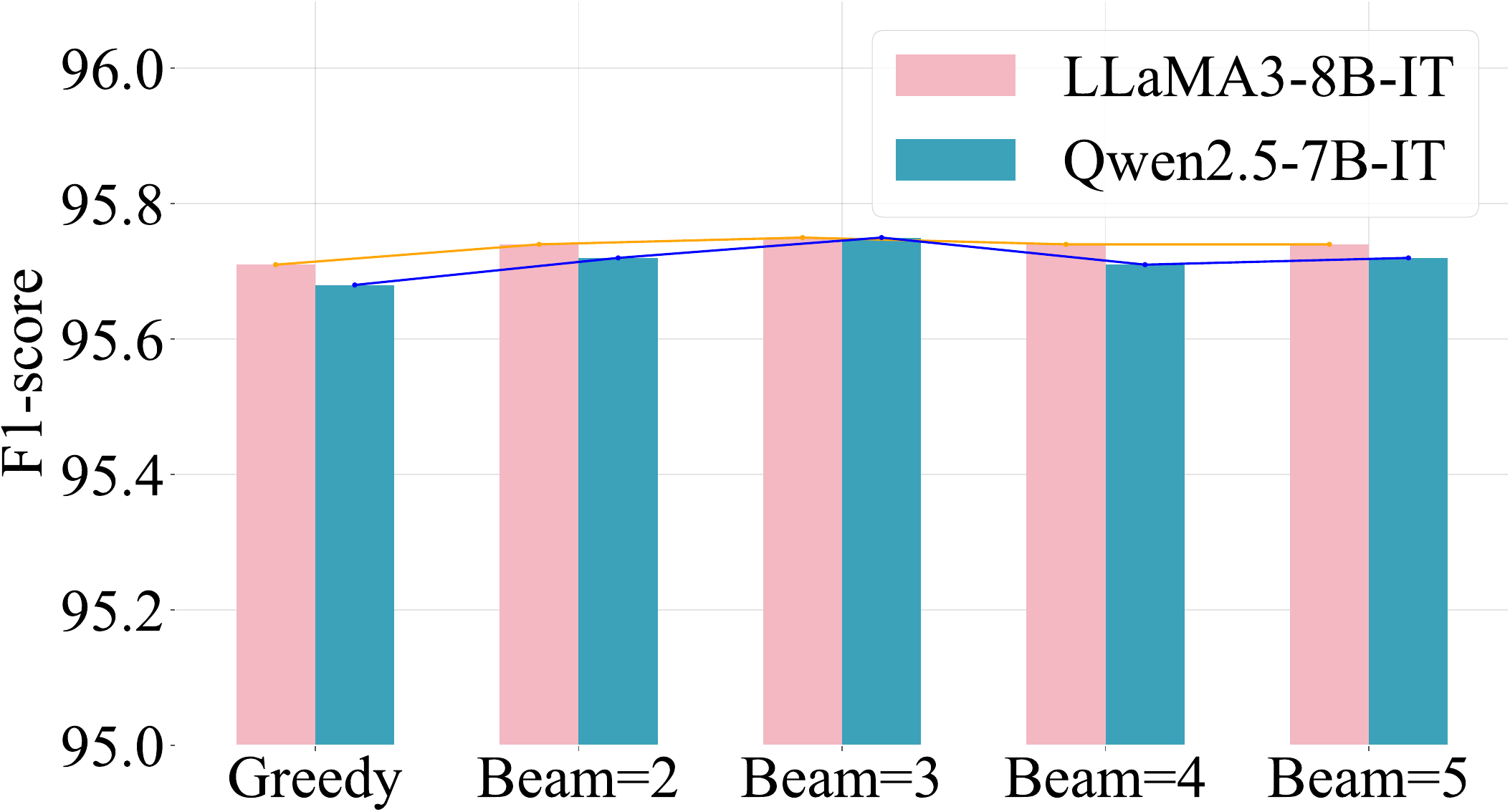}\label{fig:search}}
        \subfigure[Impact of fine-tuning strategy]{\includegraphics[width=0.3\hsize]{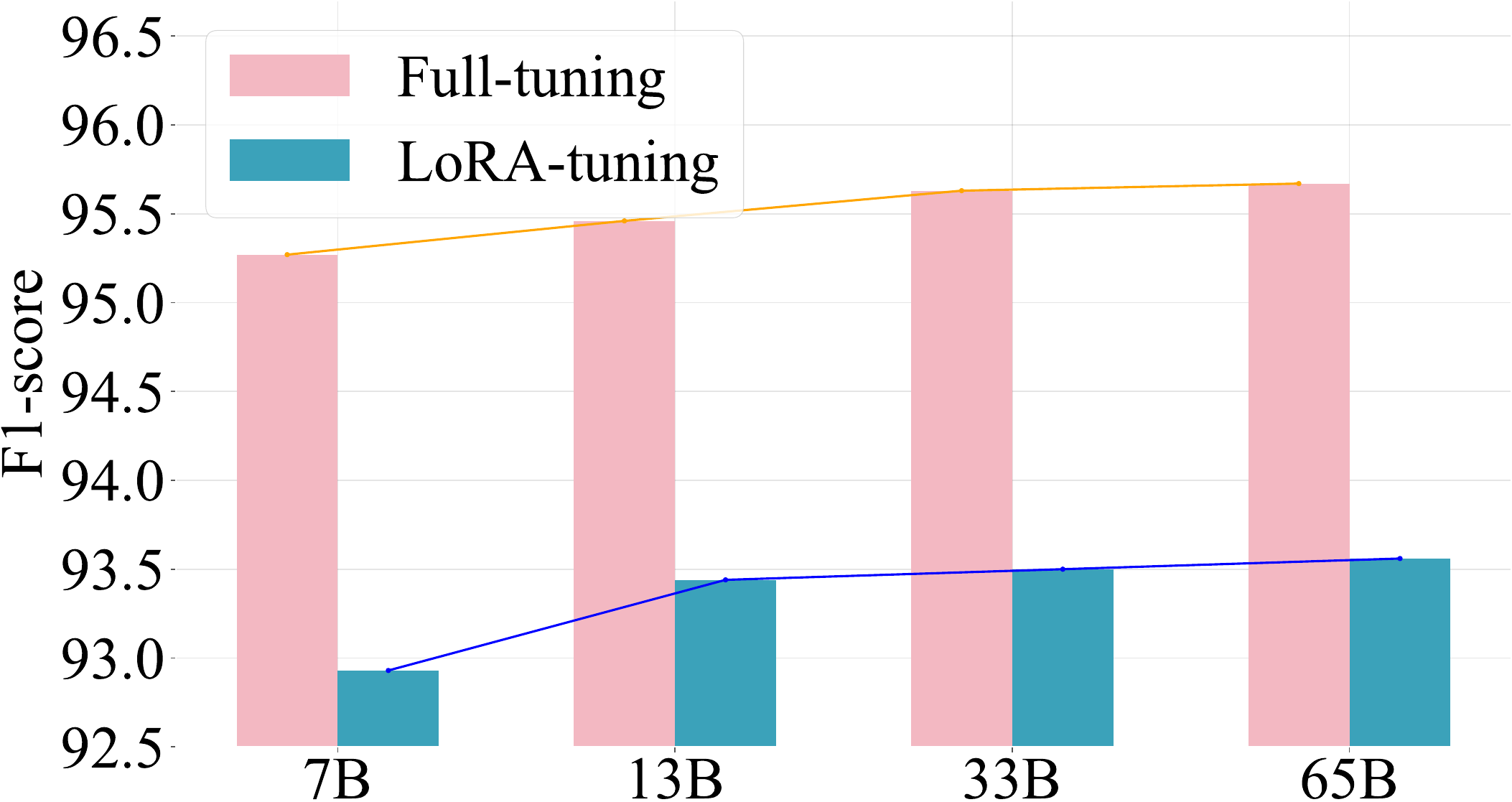}\label{fig:lora}}
	\caption{Development experiments results on the devset of PTB.}
	\label{fig:devexp}
	\vspace{-1.2em}
\end{figure*}

We first investigate which serialization format is the most preferred by LLMs for constituency parsing.
Specifically, we take LLaMA3-8B-instruct as the basic model and compare the performance of the model on the development dataset of PTB.\footnote{We have also experimented with ChatGPT, GPT-4, and LLaMA, and the overall trends are similar to LLaMA.}
Figure~\ref{fig:serilization} provides a comparison of three serialization formats under zero-shot, five-shot, and fine-tuning settings.
Under the zero-shot setting, it can be observed that LLMs initially exhibit weak capabilities in generating all three types of trees. 
This is intuitive since linearized trees are different from pre-training text and the task of consistency parsing is new to LLMs.
Under the few-shot setting, 
bracket-based linearization gives the best result among the three variants, with an F1 score of 26.44.
Span-based linearization obtains the second-best result, which is about 3 points lower than the bracket-based method.
In contrast, the performance of LLMs is poor when they are prompted to generate transition-based trees.
This suggests that transition-based trees present more challenges than bracket-based and span-based linearization methods.

In addition, bracket-based method sees better results than transition/span-based methods under the fine-tuning setting. The reason can be twofold: 1) transition-based constituent parsing requires the implicit maintenance of a stack and a buffer, thus is hard for LLMs. 2) transition-based and span-based linearization are longer than bracket-based one, thus suffering more from error propagation.
In the subsequent experiments, we opt for the bracket-based linearization strategy for comparison, as it is relatively shorter and generally delivers better performance than the transition-based method.

We then study the impact of different decoding strategies during inference.   Figure~\ref{fig:search} compares the performance of greedy search and beam search with beam sizes ranging from 2 to 5, on the development dataset of the PTB. 
Generally, the decoding strategy has a marginal impact on parsing performance, whereas beam search requires more calculations than greedy search.
Moreover, we observe a similar trend on Qwen2.5-7B-IT. \footnote{We do not report the results of LLaMA3-70B-IT and Qwen2.5-72B-IT since they exceed GPU memory limit when beam size $\geq$ 4.}. We thus adopt greedy search in subsequent experiments.

Figure~\ref{fig:lora} shows the impact of the LoRA-tuning~\cite{HuSWALWWC22} among different scales of LLaMA model. 
Compared to full-parameter tuning, LoRA-tuning yields relatively lower performance. 
For instance, the LoRA-tuned LLaMA-7B records an F1 score of 92.93, which is 2.34 points lower than full-tuned.
This observation is different from existing findings on sentence classification and text generation tasks~\cite{HuSWALWWC22,xu2023baize} which suggest LoRA-tuning can deliver results that are comparable to those achieved through fully fine-tuning. 
A potential reason could be that the output format of constituency parsing is more complex than that of other text-based tasks, thereby necessitating more substantial parameter modifications.


\begin{table}[!t]
\centering
\small
\caption{Zero-shot and few-shot results (F1) on PTB.}
\begin{tabular}{l|l|cc}
\toprule
\multicolumn{2}{c}{\textbf{Model}} & \textbf{Zero-Shot} & \textbf{Five-Shot} \\
\midrule
\multirow{2}{*}{SoTA}  & PUCP & 55.70 & \enspace - \\
  & UPCT & 62.80 & \enspace - \\
\midrule
\multirow{2}{*}{\makecell[l]{Close\\LLMs}}&ChatGPT   & 55.27 &66.41 \\
&GPT-4      & \textbf{73.00} &\textbf{73.28} \\
\midrule
\multirow{7}{*}{\makecell[l]{Open\\LLMs}}&OPT-6.7B      &\enspace0.00 &\enspace0.05  \\
&LLaMA-7B    &\enspace 0.00 & \enspace 8.82 \\
&Vicuna-7B    &\enspace 0.94 & 12.17 \\
&Qwen2.5-7B-IT  &27.80  &33.01  \\
&LLaMA3-8B-IT   &20.30  &26.59  \\
&LLaMA3-70B-IT  &44.97 &60.41 \\
&Qwen2.5-72B-IT  &57.99 &62.13 \\

\bottomrule
\end{tabular}
\label{table:fewshot}
\end{table}

\subsubsection{\textbf{Are LLMs Still Zero/Few-shot Learners for Constituency Parsing?}}
Previous work has shown that LLMs have strong generalization abilities across tasks and domains~\cite{wei2022finetuned,openagi,zhou2023lima}.
To study this, we evaluate the generalization ability of LLMs under both in-domain and cross-domain settings. 
Table~\ref{table:fewshot} documents the zero-shot and 5-shot learning performance of different LLMs on PTB.\footnote{Due to the difficulty of conducting large-scale evaluations with GPT-4, we conduct evaluations on a subset of 500 samples, which is randomly selected from the original testsets.} 
It is noteworthy that commercial LLMs (ChatGPT and GPT-4) deliver impressive results in both scenarios, achieving approximately 73 in F1 scores under both zero-shot and 5-shot settings.
In contrast, open-source LLMs show relatively lower performance. 
Specifically, OPT-6.7B and LLaMA-7B deliver low results under the zero-shot setting and give less than 10 F1 scores under the five-shot setting.
Compared with them, Vicuna-7B, Qwen2.5-7B-IT, LLaMA3-8B-IT, LLaMA3-70B-IT, and Qwen2.5-72B-IT achieves better results. 
A possible reason is that these models are trained on instruction-tuning datasets, which might help LLMs to shape the output according to instructions.

\begin{table}[!t]
\centering
\small
\caption{Fine-tuning results on PTB. LR: labeled recall. LP: labeled precision. The best results among all methods are \textbf{bolded} and the best sequence-based results are \underline{underlined}.}
\begin{tabular}{l|l|cccccc}
\toprule
\textbf{Category}&{\textbf{Model}}& \textbf{LR}& \textbf{LP}& \textbf{F1}   \\
\midrule
\multirow{2}{*}{\makecell{Graph}}&SEPar   & 95.56 & 95.89 & 95.72 \\
 &SAPar &\textbf{96.19} & \textbf{96.61} & \textbf{96.40}        \\
 \midrule
 Transition&TGCN  & 96.13 & 96.55 & 96.34        \\
 \midrule
 \multirow{3}{*}{\makecell{Sequence\\w/o LLM}}&LSTM   & - & - &88.30    \\ &Transformer   &- &- &91.20     \\
 &GPT-2  &93.68 &93.79 & 93.73    \\
\midrule
\multirow{7}{*}{\makecell{Sequence\\with LLM}} &OPT-6.7B &94.63 &94.52 &   94.58  \\
 &LLaMA-7B &95.50 & 95.12&   95.31  \\
 &Vicuna-7B &95.37 & 94.93 &  95.16 \\
 &Qwen2.5-7B-IT &95.96 & 95.51 & 95.74  \\
 &LLaMA3-8B-IT  & 96.04 &95.50 & 95.77 \\
 &LLaMA3-70B-IT  &{96.07} & {95.69} &  {95.88}     \\
 &Qwen2.5-72B-IT  &\underline{96.11} & \underline{95.97} &  \underline{96.04}     \\
\bottomrule
\end{tabular}
\label{table:finetune}
\end{table}

Additionally, larger LLMs generally give better performance compared with small ones, suggesting a positive correlation between constituency parsing ability and model size. 
While LLMs may not excel as few-shot learners in the domain of constituency parsing, they do provide novel pathways for advancements in unsupervised constituency parsing.

\subsubsection{\textbf{Exploring Full Capacity of LLMs for Constituency Parsing}}
To study the full potential of LLMs on constituency parsing, we fine-tune LLMs on the full training dataset of PTB and compare the performance of LLMs with the state-of-the-art method.
Table~\ref{table:finetune} presents the results of different systems on the PTB test set. Among all LLM-based methods, Qwen2.5-72B-IT achieves the highest result, with an F1 score of 96.04.
Compared with sequence-based baselines, fine-tuned LLMs yield greatly superior results. 
Notably, Qwen2.5-72B-IT gives an improvement of approximately 2.3 F1 score over GPT-2. 
This observation suggests that LLMs can substantially boost the performance of sequence-based constituency parsers.

Compared with the state-of-the-art graph-based and transition-based systems, Qwen2.5-72B-IT gives competitive results. 
Specifically, Qwen2.5-72B-IT outperforms the SEPar parser's 95.72 F1 score, narrowing the gap between the sequence-based parser and chat-based parser.
This indicates that the LLM-based parser can serve as a strong backbone for constituency parsing.
Qwen2.5-72B-IT produces relatively lower results compared with SAPar and TGCN. {Potential reasons for this observation include: 1) SAPar and TGCN employ external mechanisms which are specifically designed for constituency parsing that are adept at capturing local features; 2) Sequence-based parsers learn syntax indirectly through language modeling objectives, prioritizing next-token prediction over structural accuracy, which leads to poor handling of long-range dependencies in complex sentences.}

\begin{table*}[t]
\centering
\small
\caption{Results on cross-domain test datasets. $\Delta$F1: relative reduction rate of F1, lower is better. Liter. refers to the Literature domain.
The best results in each group are \textbf{bolded}.}
\begin{tabular}{l|c|cccccc|c}
\toprule
\multirow{2}{*}{\textbf{Model}} & \textbf{In-domain} & \multicolumn{6}{c}{\textbf{Out-of-domains}} & \\
 &\textbf{PTB} &\textbf{Dialogue} & \textbf{Forum} & \textbf{Law}   & \textbf{Literature} & \textbf{Review} & \textbf{Avg} & \textbf{$\Delta$F1}\\ \midrule
\multicolumn{9}{l}{\textit{\textbf{Zero-shot Learning}}} \\
ChatGPT  & 55.27&  50.73  &   44.01    &    42.61   &  42.66 &49.32  & 45.87 &-\\
GPT4  & \textbf{73.00} & \textbf{65.74} &  \textbf{61.02} &  \textbf{70.39} &  \textbf{60.34}  &  \textbf{63.31} & \textbf{64.16} & -\\
LLaMA3-70B-IT  & 44.97 & 52.20 & 48.47 & 44.08 &  44.77  & 48.12 & 47.53 & -\\
Qwen2.5-72B-IT & 57.99 & 55.83 & 53.14 & 53.59 & 51.28 & 53.14 & 53.40 &- \\
\midrule
\multicolumn{9}{l}{\textit{\textbf{Five-shot Learning}}} \\
ChatGPT   & 66.41 & 61.40 &  56.19   &  56.33 &  51.18   &   59.08&  56.84 &14.41\% \\
GPT4   & \textbf{73.28} & \textbf{66.61} &  \textbf{62.01} & \textbf{72.43} & \textbf{58.29} &  \textbf{66.41} & \textbf{65.15}  &{11.09}\%\\
LLaMA3-70B-IT  & 60.41 & 60.31 &  55.54 & 58.95 &  53.41  &  57.79 & 57.20 & \enspace\textbf{5.31}\%\\
Qwen2.5-72B-IT  & 62.13 & 60.23 & 56.72 & 60.38 &  52.49  &  58.43 & 57.65 & \enspace7.21\%\\
\midrule
\multicolumn{9}{l}{\textit{\textbf{Supervised Fine-tuning}}} \\
SEPar$\texttt{[Graph]}$ & 95.72 & \textbf{86.30}  & \textbf{87.04} &  \textbf{92.06}  &  \textbf{86.26} & \textbf{84.34} & \textbf{87.20} & \textbf{\enspace8.90\%} \\
SAPar$\texttt{[Graph]}$ & \textbf{96.40} & 86.01  & 86.17 &  91.71  &  85.27 & 83.41 &  86.51& 10.26\% \\
GPT-2 & 93.73 & 80.60 & 79.42 & 83.10  & 74.87 & 77.09 & 79.02 &15.69\% \\
LLaMA-7B  & 95.31 & 82.92  &  81.56 &  84.92  &    79.19  &  78.86 & 81.49&14.50\% \\
Qwen2.5-7B-IT  & 95.74  &  84.13 &    81.35   &  85.47   &  79.56          &   79.72  & 82.05 &14.30\%\\
LLaMA3-8B-IT  & 95.77  &  83.79 &     81.65   &  84.99   &  79.81          &   79.75  & 82.00 &14.38\%\\
LLaMA3-70B-IT & 95.88 &  {84.21} & 82.48  &   85.54    &   79.63  & 80.90 &  82.55 & {13.90\%}\\
Qwen2.5-72B-IT  &{96.04}   &{84.20}   &{82.64}   &{85.73}      &{79.76}    &{81.41} &{82.75} &{13.84\%}     \\
\bottomrule
\end{tabular}
\label{table:cross-domain}
\end{table*}

\subsubsection{\textbf{Generalization across Domains}}

To study the cross-domain generalization capability of LLMs, we evaluate the unsupervised domain adaptation performance in five out-of-domain test sets.
Table~\ref{table:cross-domain} shows the F1 score on different datasets and the relative performance reduction rate. 
Remarkably, ChatGPT, GPT-4, LLaMA3-70B-IT, and Qwen2.5-72B-IT obtain impressive performance without training.
Compared with GPT-2, the Fine-tuned LLMs (including LLaMAs and Qwens) show relatively lower performance reduction rates, and these reduction rates decrease as the model parameters increase.
This indicates that LLMs exhibit superior generalization capability in the context of domain shifting when compared to earlier generatively pre-trained models.

Moreover, the relative performance reduction rates of fine-tuned sequence-based parsers (including LLaMAs and GPT-2) are larger compared to SEPar and SAPar. 
This shows, somewhat surprisingly, that LLMs-based parsers are weaker than traditional chart-based parsers in cross-domain parsing, despite pre-training on massive textual data.
The reason can be that sequence-based parsing paradigm tends to learn complex sequence-to-sequence mapping, making it vulnerable to changes at the span level. 
Consequently, this limits its ability to generalize effectively across different domains. {Future studies may explore domain adaptation techniques (e.g., targeted fine-tuning~\cite{GururanganMSLBD20}, model merging~\cite{lu2025fine}) to mitigate cross-domain performance degradation}.


\begin{figure}[!t]
	\setlength{\abovecaptionskip}{0.03cm}   
	\setlength{\belowcaptionskip}{-0.1cm}
	\centering 
	\subfigure[Impact of input length]{\includegraphics[width=0.47\hsize]{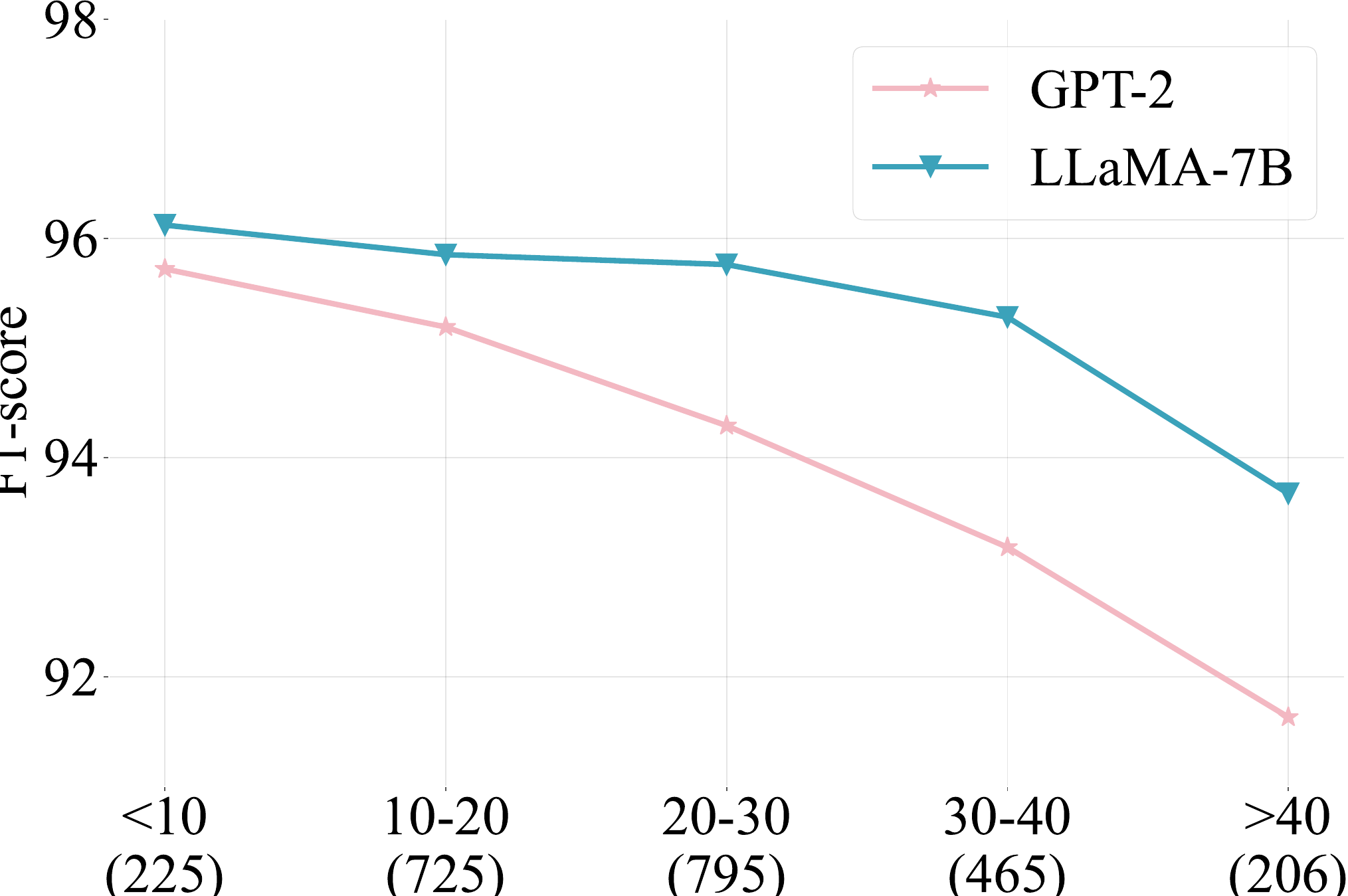}\label{fig:length1}}
	\subfigure[Impact of span length]{\includegraphics[width=0.50\hsize]{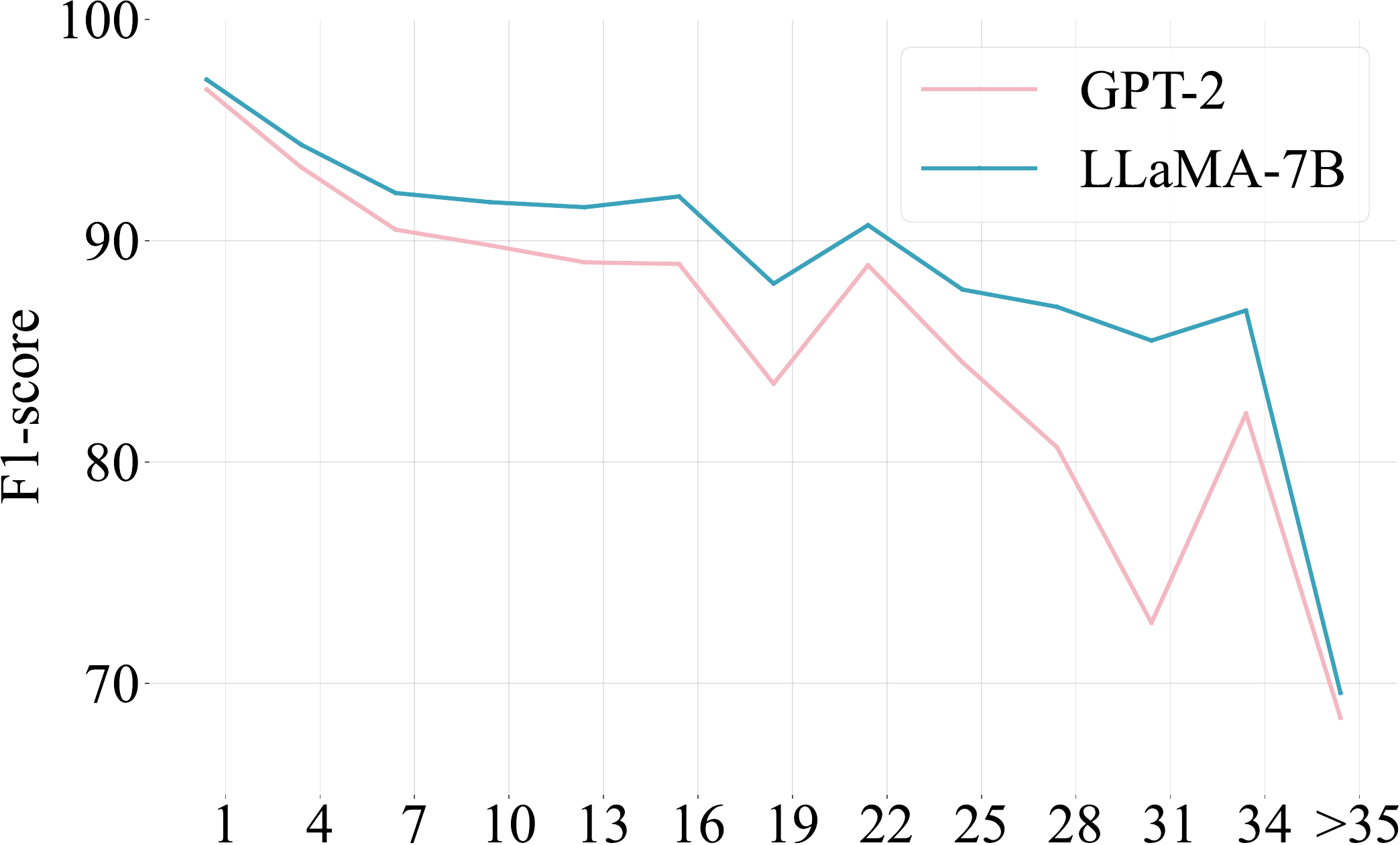}\label{fig:length2}}
	\caption{Performance across varying input lengths and constituent spans.}
	\label{fig:lenexp}
\end{figure}

\subsubsection{\textbf{Generalization across Lengths}}

Since LLMs are trained with a longer context, they are expected to have a better ability on long inputs.
To study this, we partition the test set into three subsets based on sentence length: $\leq$10, 10-20, 20-30, 30-40, $\geq$40, and evaluate the model performance. \footnote{{We also tried partitioning the test set according to different length intervals, the overall trend is similar.}}
We compare an LLM-based parser (LLaMA-7B) with a sequence-based baseline (GPT-2). 
Figure~\ref{fig:length1} shows the performance of tree models against input length.
All three models give relatively lower results when parsing long input sentences.
GPT-2 shows the largest performance reduction among the three models.
Compared with GPT-2, LLaMA-7B sees lower performance decreases on long sentences, showing that LLMs are more robust to longer sentences. 
When input sentences have more than 40 tokens, both models exhibit diminished generalization capabilities.
This indicates that LLMs do not offer a definitive solution to fundamentally address the performance limitation of sequence-based parsers.


Since LLMs are trained with longer context than traditional models, they are expected to have a better ability to predict long spans.
To verify this, we calculate the F1-score based on span lengths and compare the performance of LLM-based parser and the traditional sequence-based parser.
Figure~\ref{fig:length2} shows the F1-score of Llama-7B and GPT-2 on constituent spans with different lengths.
Generally, LLaMA-7B gives better results than GPT-2 in most span length conditions.
Moreover, LLaMA-7B shows larger improvements on medium and long spans, suggesting that LLMs-based parser has a better ability to learn more sophisticated constituency label dependencies.
Interestingly, both models show great performance drops when the span length is greater than 35, suggesting that generative models still have limitations in handling extremely long spans.
This limitation may be attributed to the fact that sequence-based parsers are naturally weak in modeling long-range nested dependencies.

\begin{table*}[!t]
\centering
\small
\caption{Valid F1/Overall F1/Invalid rate of different models on in-domain and out-of-domain testsets. ZS: zero-shot learning, FS: five-shot learning, and FT: fine-tuning.}
\begin{tabular}{lcccc}
\toprule
\textbf{Domain} & \textbf{SEPar} & \textbf{Qwen2.5-72B-IT (ZS)} & \textbf{Qwen2.5-72B-IT (FS)} & \textbf{LLaMA-7B (FT)} \\ 
\midrule
PTB &  95.72/95.72/0.00 &  72.56/57.99/41.58 & 74.78/62.13/30.46 & 95.43/95.31/0.50  \\
Dialogue &  86.30/86.30/0.00 &67.49/55.83/30.90 & 68.63/60.23/23.40 & 84.26/82.92/4.00  \\
Forum &  87.04/87.04/0.00 &67.24/53.14/43.30 & 68.87/56.72/38.70 &  84.08/82.56/9.20    \\
Law &  92.06/92.06/0.00 &73.42/53.59/50.40 &78.34/60.38/44.50 & 90.69/84.92/16.20   \\
Literature &  86.26/86.26/0.00  &68.61/51.28/51.10 &72.34/52.49/50.80 & 82.94/79.19/16.10  \\
Review &  84.34/84.34/0.00 &64.48/53.14/27.80 &69.31/58.43/18.90 & 81.54/78.86/4.60  \\
\hline
\end{tabular}
\label{tab:hurt}
\end{table*}

\section{Error Analysis}
{Different from chart-based models, a drawback of sequence-based models, including LLMs, is that there is no mechanism, such as the CKY algorithm~\cite{pereira1983parsing} or using a state buffer~\cite{yamada-matsumoto-2003-statistical}, which can ensure the validity and faithfulness of the generated constituency parse trees.
On the other hand, due to the token-by-token generation nature, LLMs face challenges in maintaining global validity during local token prediction, often hallucinating unfaithful contents when the model prioritizes local coherence over structural faithfulness.}
In this section, we conduct an error analysis to study whether this impacts the performance of LLMs.
For the zero-shot and few-shot learning settings, we select Qwen2.5-72B-IT, whereas for the supervised fine-tuning setting, we utilize the LLaMA-7B. 
Our analysis focuses on the parsing results of these LLMs across both in-domain and out-of-domain test sets.

\subsection{LLMs Can Fail to Output Valid Trees}
We first study whether LLMs generate invalid trees. To this end, we calculate the rate of invalid tree generation for the fine-tuned LLaMA-7B, Qwen2.5-72B-IT under the zero-shot setting, and Qwen2.5-72B-IT under the five-shot setting, across six test sets.
To further study whether invalid trees significantly affect the model performance, we calculate the F1-score for valid trees and compare it with the overall F1-score.
As shown in Table~\ref{tab:hurt}, the invalid rates of SEPar are consistently zero in all testsets. 
In contrast, the invalid rates of LLMs are relatively higher compared with SEPar, especially under zero-shot and five-shot settings. 
Notably, Qwen2.5-72B-IT generates more than 50\% invalid trees on the literature domain under the zero-shot learning setting.
In addition, invalid trees greatly hurt the overall performance of LLM-based parsers. 
In particular, the fine-tuned LLaMA-7B obtains an F1 score of 90.69 on the law domain for valid trees, which is about 6 points higher than the overall performance, even with fine-tuning.
This explains why the fine-tuned LLMs achieve significantly lower results on out-of-domain datasets.

To further understand the problem of invalid tree generation, we divide invalid trees into different groups and analyze the distribution of each group.
We categorize invalid trees into four types: more than one word, missing words, bracket unmatched, and other.
``more than one word'' refers to instances where a single constituency tree leaf contains multiple words;
``missing word'' refers to instances where a single constituency tree leaf does not contain any words;
``bracket unmatch'' refers to instances where the number of left and right brackets are not matched;
``Other'' refers to other cases, such as the model totally fails to predict a tree.
Figure~\ref{fig:error} shows the distribution of invalid trees generated by the fine-tuned LLaMA-7B on all out-of-domain testsets. 
It can be observed that ``more than one word'' and ``missing word'' are the two most frequent errors, accounting for 35$\%$ and 32$\%$, respectively. ``bracket unmatch'' is also a frequent error, particularly prevalent in lengthy sentences or those containing special tokens.



\begin{figure}[!t]
    \centering
    \includegraphics[width=0.37\textwidth]{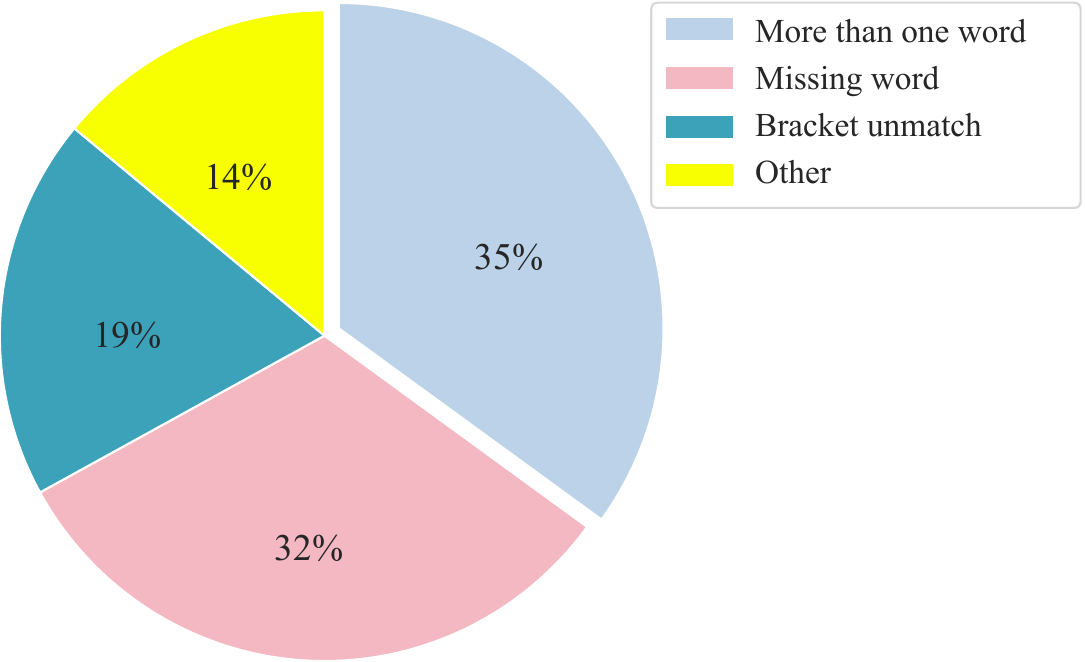}
    \caption{Distribution of invalid trees generated by LLaMA-7B.}
    \label{fig:error}
\end{figure}

\begin{table}[!t]
\centering
\small
\caption{The average unfaithful tree rate of different models across all datasets. OverGen: over generation, WordMis: word mismatch, Failure: prediction failure.
}
\begin{tabular}{l|ccc}
\toprule
 \textbf{Model} & \textbf{OverGen}    & \textbf{WordMis} & \textbf{Failure}\\ 
\midrule
Qwen2.5-72B-IT (ZS)  &\textbf{27.30}  &5.60  &\textbf{8.11} \\ 
Qwen2.5-72B-IT (FS)  &25.12  &\textbf{6.02}  & 3.79  \\ 
LLaMA-7B (FT)  &\enspace5.15  &1.80  &0.22  \\
\bottomrule
\end{tabular}
\label{table:hallucination}
\end{table}

\subsection{LLMs Can Hallucinate Unfaithful Constituents}
Hallucination~\cite{maynez-etal-2020-faithfulness,bang2023multitask} is still an unsolved problem for LLMs, which has been observed in many generation tasks.
In constituency parsing, we find that LLM-based parsers frequently generate constituency trees that are unfaithful to input sentences. 
To further study this, we categorize the ``\textit{hallucinated}'' trees into three types: over generation, word mismatch, and prediction failure.
``\textit{over-generation}'' refers to a situation where the number of words in the predicted tree differs from the original text;
``\textit{word mismatch}'' refers to a situation where a word in the predicted constituent parse tree has been modified, resulting in a mismatch with the original text\footnote{{For ``\textit{word mismatch}'' error, the predicted sentence has the same length as the input sentence, while the correctness of the tree structure is not guaranteed.}};
``\textit{prediction failure}'' refers to a scenario where the model totally fails to perform the constituency parsing task.

Table~\ref{table:hallucination} presents the averaged percentage of hallucination trees on all testsets.
We choose Qwen2.5-72B-IT for analysis in zero-shot and five-shot learning settings and LLaMA-7B for supervised learning settings. 
Among all categories, the most frequent error is the over-generation error, accounting for about 70$\%$ of the total number. 
We further categorize length mismatch errors into three subcategories:
\textit{repetition}, \textit{continue writing}, and \textit{other errors}.
Among them, ``\textit{repetition}'' is frequently observed in generative models, which refers to the phenomenon where model outputs contain large repeated contents~\cite{holtzman2019curious}.
In addition, word mismatch accounted for about 15\%. LLMs frequently alter numerical values, nouns, and adjectives within input sentences, such as changing ``84'' to ``81''.
Furthermore, prediction failures account for about 15\% of hallucinated trees.
Finally, it can be observed that the number of hallucination errors is larger under zero-shot and five-shot learning settings compared with supervised fine-tuning. 
This is intuitive since LLMs are not sufficiently trained under these settings.

\begin{figure*}[!t]
	\centering
	\includegraphics[width=0.90\textwidth]{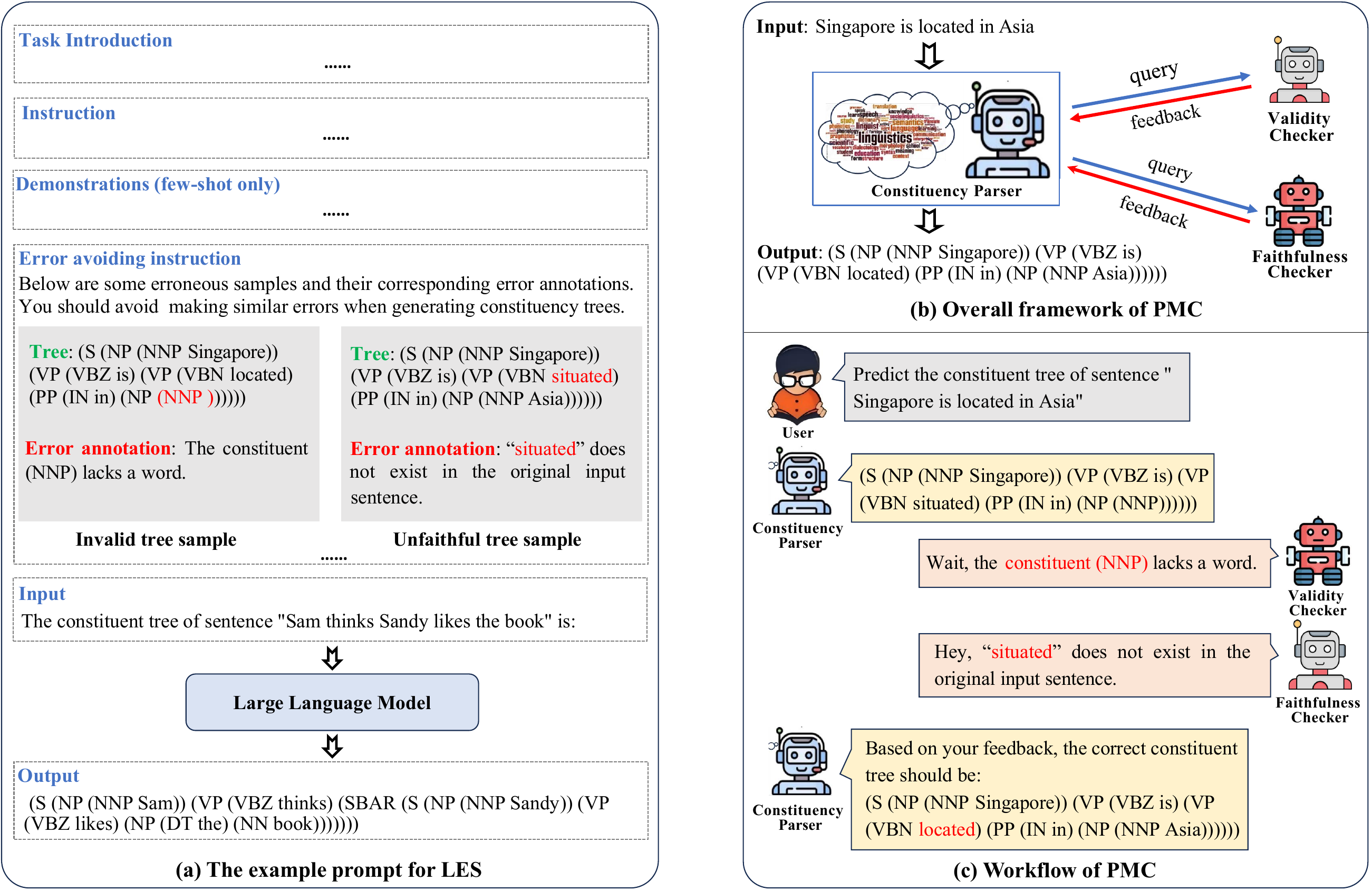}
	\caption{Illustration of the proposed methods.}
	\label{fig:our-method}
\end{figure*}

\section{Mitigating Invalid and Unfaithful Constituent Trees}
Based on the observation that LLMs frequently produce invalid and hallucinated trees in constituency parsing, we propose two strategies to enhance the validity and faithfulness of the generated constituent trees. 
The first strategy is \textit{learning from erroneous samples}, which guides LLMs to gain insights about how to avoid generating invalid or unfaithful trees by incorporating erroneous samples into prompts.  
The second strategy is \textit{parsing with multi-agent collaboration}, which employs two additional agents to provide feedback on the validity and faithfulness of the initially generated tree and then use this feedback to refine the constituency tree.
{LES and PMC are parallel parsing strategies designed for distinct use cases: LES suits latency-sensitive applications since it requires a single forward pass to generate trees; PMC relies on multi-agent interaction, significantly increasing the computation cost. Therefore, PMC favors accuracy-critical scenarios.\footnote{{We also tried to combine two strategies, but did not observe significant improvements.}}}
We employ the \textit{learning from erroneous samples} strategy under zero-shot and few-shot settings, and the \textit{parsing with multi-agent collaboration} strategy can be applied in zero-shot, few-shot, and supervised fine-tuning settings.

\subsection{Learning from Erroneous Samples}
{It is well-documented that human has a strong cognitive ability to rapidly learn from negative examples - a phenomenon observed in both psychological studies and educational research~\cite{Kornell2009UnsuccessfulRA,PMID:27648988}. 
Inspired by this, we introduce the \textit{learning from erroneous samples} (LES) strategy to guide LLMs to avoid producing invalid or unfaithful trees by leveraging insights gained from erroneous samples and their corresponding annotations, therefore improving LLMs under the zero-shot and few-shot settings.} 
As shown in Figure~\ref{fig:our-method}(a), we leverage erroneous samples as well as their error annotation to construct an \textit{error-avoiding instruction}, before prompting LLMs for constituency parsing. 
In this way, we expect LLMs to gain insights from erroneous samples, thus avoiding generating erroneous trees.
We consider the following two types of erroneous samples and their annotation to an \textit{error-avoiding instruction}.

\noindent\textbf{Invalid samples and corresponding annotations}. We add several trees that contain different invalid constituents to the input prompt of LLMs, such as ``\textit{(S (NP (NNP Singapore)) (VP (VBZ is) (VP (VBN located) (PP (IN in) (NP \textcolor{red}{(NNP )})))))}''. 
In addition, we manually write error annotations and append them to the input, such as ``\textit{The constituent (NNP) lacks a word.}'' 

\noindent\textbf{Unfaithful samples and corresponding annotations}. We also add several trees that contain unfaithful constituents to the input prompt, such as ``\textit{(S (NP (NNP Singapore)) (VP (VBZ is) (VP (VBN \textcolor{red}{situated}) (PP (IN in) (NP (NNP Asia))))))}''. 
In addition, we manually write error annotations and append them to the input, such as ``\textit{`situated' does not exist in the original input sentence.}'' 

{We manually curate four examples\footnote{{The manual annotation process required approximately 40 minutes in total to construct 8 demonstration samples.}} to construct an targeted \textit{error-avoiding instruction}, which is subsequently appended to the base instruction depicted in Figure~\ref{fig:prompt} to guide LLMs in performing constituency parsing.}\footnote{{We also tried to use other four groups of examples, with the results showing comparable performance to those reported in our main experiments.}}

\subsection{Parsing with Multi-agent Collaboration}
We propose a new framework, named \textit{parsing with multi-agent collaboration} (PMC), that reduces the generation of invalid and unfaithful constituent trees through multi-agent collaboration. 
As depicted in Figure~\ref{fig:our-method}(b), our framework comprises three distinct agents: a constituency parser, a validation checker, and a faithfulness checker. 
The constituency parser, endowed with linguistic expertise, aims to predict the constituency tree for the provided input.
The validation checker and faithfulness checker serve as evaluative agents, delivering feedback focused on validation and faithfulness, respectively, to improve the precision of the constituent tree produced by the parser. 
As depicted in Figure~\ref{fig:our-method}(c),
this framework works in an iterative way, beginning with the constituency parser generating an initial tree. Subsequently, the validation checker and faithfulness checker provide feedback based on the generated tree. 
Finally, the constituency parser refines the tree using both sets of feedback in conjunction with the initial tree.

\noindent\textbf{Constituency Parser:} It aims to generate a constituent tree based on the input sentence. 
To prevent the production of invalid or unfaithful trees, it additionally integrates feedback from the two tree checkers into its input prompt to guide LLMs. 
To ensure consistency between the inputs of the first and subsequent rounds, the constituency parser is fed with empty feedback during its initial parsing round.

\noindent\textbf{Validation Checker:} It is tasked with assessing the structural integrity of the trees produced by the 
constituency parser, and giving corresponding feedback to the parser. Under the supervised setting, we first construct invalid trees with rules based on the training dataset, then fine-tune an LLM to automatically evaluate the tree and give corresponding feedback. 
Under the zero-shot or few-shot setting, we employ the same rules to construct task instructions and demonstrations to guide an LLM to achieve this goal.

\noindent\textbf{Faithfulness Checker:} It is designed to assess and give feedback on the semantic fidelity of the trees, with a primary focus on evaluating the alignment between the words in the constituent tree and those in the input sentence. 
Under the supervised setting, we first construct unfaithful trees by replacing words with their synonyms, then fine-tune an LLM to identify these substitutions and provide revision suggestions.
Under the zero-shot or few-shot setting, we employ a task instruction to prompt an LLM to act as the faithfulness checker.

\begin{figure}[!t]
	\setlength{\abovecaptionskip}{0.03cm}   
	\setlength{\belowcaptionskip}{-0.1cm}
	\centering 
	\subfigure[zero-shot]{\includegraphics[width=0.32\hsize]{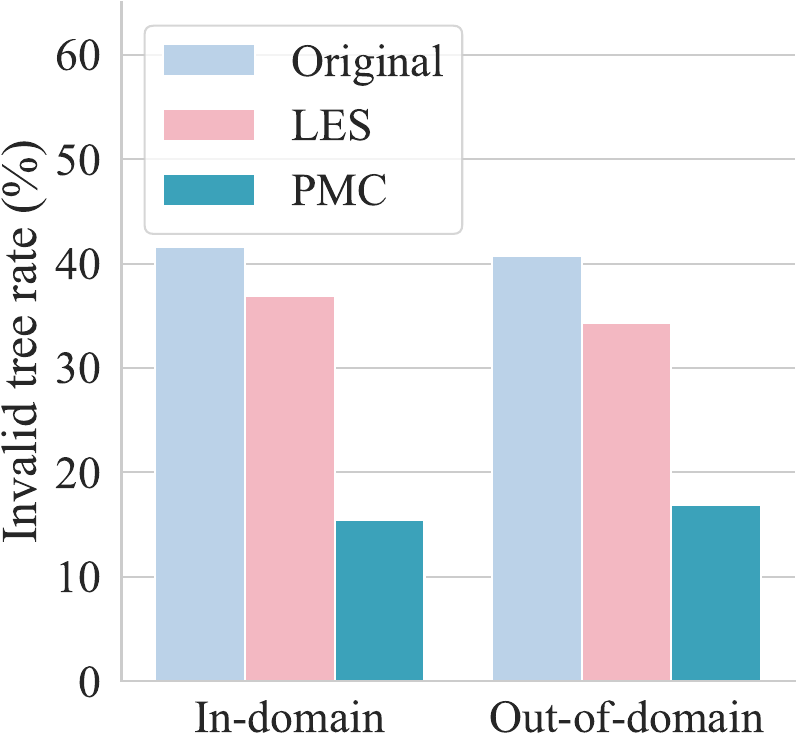}\label{fig:rate1}}
	\subfigure[five-shot]{\includegraphics[width=0.32\hsize]{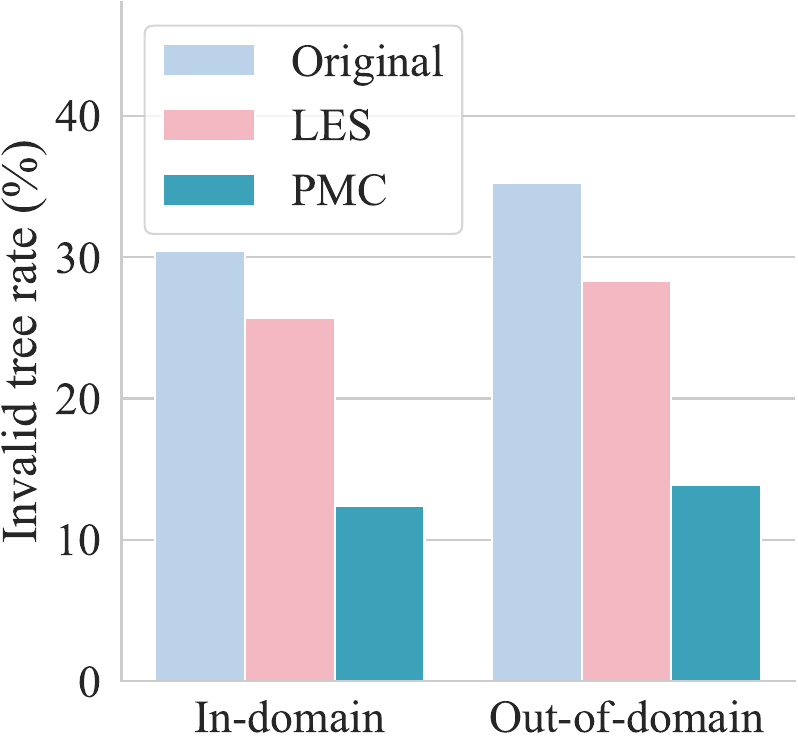}\label{fig:rate2}}
        \subfigure[fine-tuning]{\includegraphics[width=0.32\hsize]{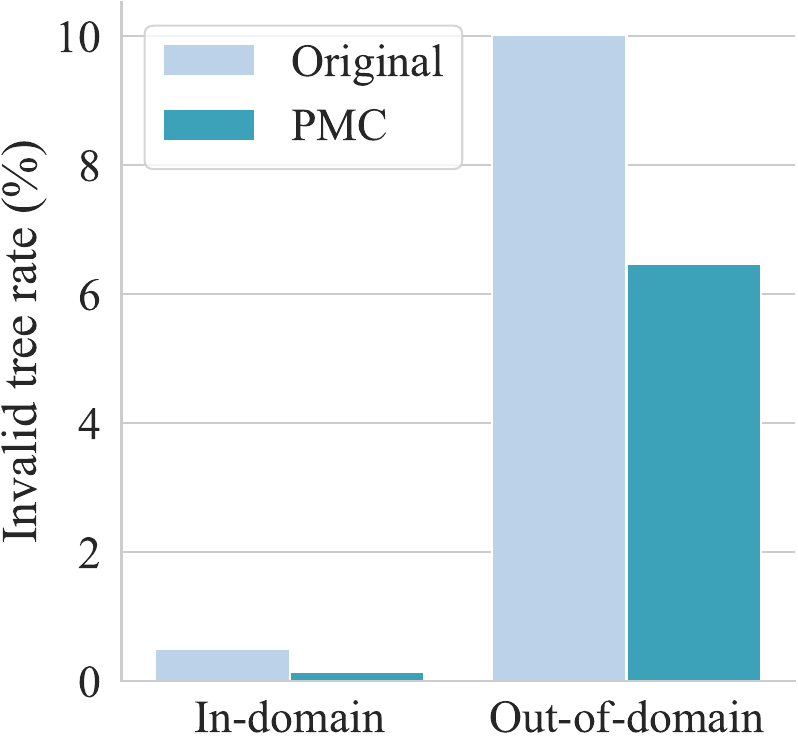}\label{fig:rate3}}
	\caption{Invalid tree rate of different systems.}
	\label{fig:rate-valid}
\end{figure}

\begin{figure}[!t]
	\setlength{\abovecaptionskip}{0.03cm}   
	\setlength{\belowcaptionskip}{-0.1cm}
	\centering 
	\subfigure[zero-shot]{\includegraphics[width=0.32\hsize]{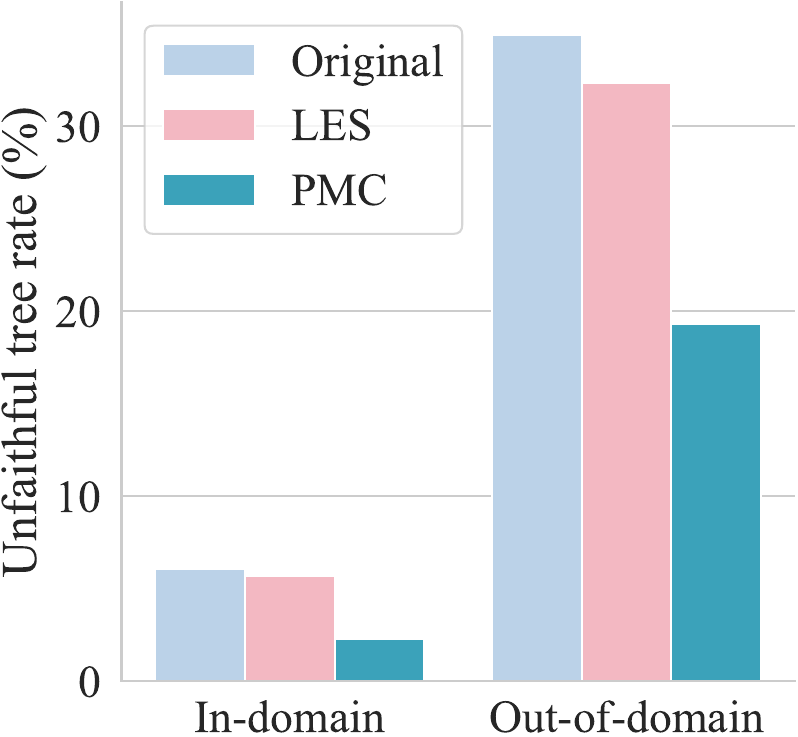}\label{fig:rate21}}
	\subfigure[five-shot]{\includegraphics[width=0.32\hsize]{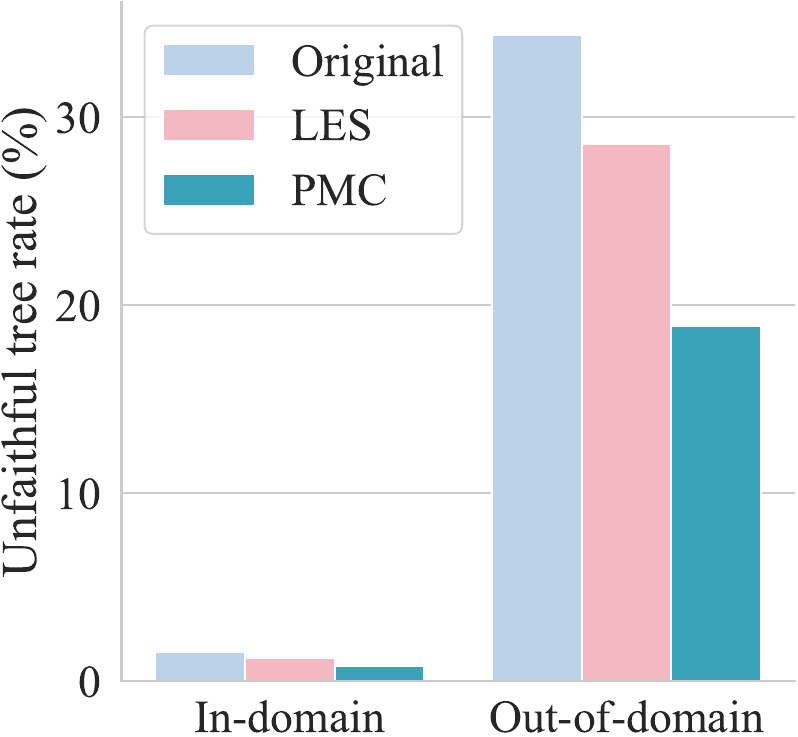}\label{fig:rate22}}
        \subfigure[fine-tuning]{\includegraphics[width=0.32\hsize]{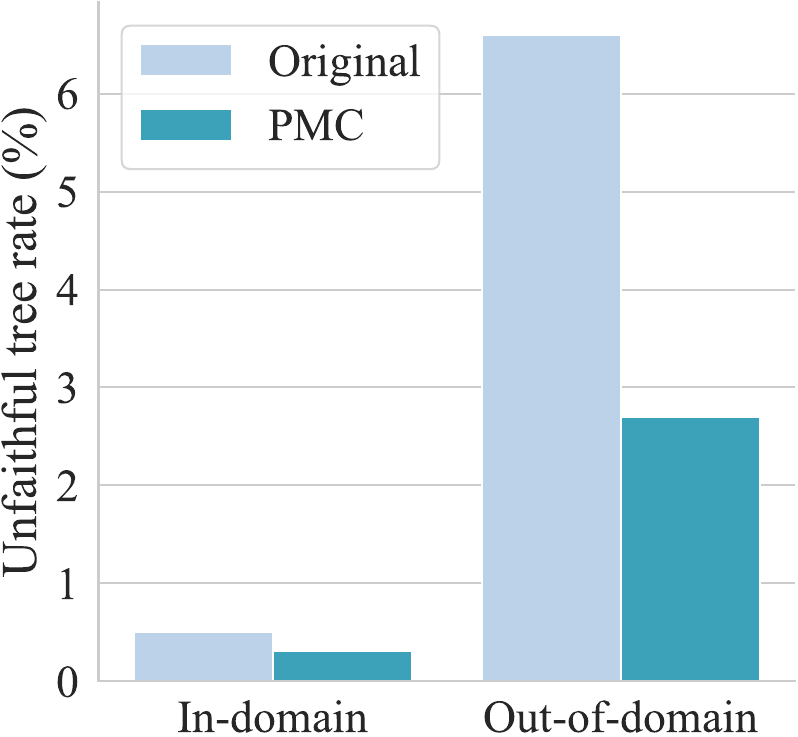}\label{fig:rate23}}
	\caption{Unfaithful tree rate of different systems.}
	\label{fig:rate-faithful}
\end{figure}

\subsection{Do LES and PMC Reduce Invalid and Unfaithful Constituent Trees?}
Figure~\ref{fig:rate-valid} illustrates the rate of invalid trees generated by different systems when evaluated under zero-shot, five-shot, and supervised fine-tuning settings. Results are shown for both in-domain (PTB) and out-of-domain datasets (averaged scores across five test sets). 
As depicted in Figure~\ref{fig:rate-valid}, both LES and PMC consistently reduce the rate of invalid trees across all datasets. 
PMC consistently outperforms LES on all datasets, likely due to its capacity to provide fine-grained feedback rather than general instructions, thus enabling more precise error correction. 
Under the supervised learning setting, PMC shows greater improvements on out-of-domain datasets compared to in-domain datasets. 
This is because the number of invalid and unfaithful trees is significantly higher in out-of-domain datasets, as supported by the findings in Table~\ref{tab:hurt}.

Figure~\ref{fig:rate-faithful} shows the unfaithful tree rate across various models. 
Overall, both LES and PMC demonstrate a reduction in the unfaithful tree rate across all datasets, confirming the efficacy of these two methods in mitigating the production of both invalid and unfaithful constituent trees.
Similar to the trends observed in Figure~\ref{fig:rate-valid}, PMC achieves relatively superior outcomes compared to LES.

\subsection{Do LES and PMC Improve the Overall Performance?}
\label{sec:our-res}

To evaluate the effectiveness of the proposed methods, we implemented them on top-performing models, including GPT4, LLaMA3-70B-IT, and Qwen2.5-72B-IT. 
We assessed their performance on both in-domain and out-of-domain test sets, comparing them with SoTA approaches. 
Table~\ref{tab:new-results-zs} presents a comparison of the F1-scores for different systems on five datasets across zero-shot, five-shot, and supervised fine-tuning learning settings\footnote{We did not experiment with GPT4 under the supervised fine-tuning setting, as we do not have access to its checkpoint.}.

Firstly, it can be observed that both LES and PMC consistently enhance the performance of the two baseline systems under the zero-shot learning setting. 
Notably, LES and PMC achieve average F1 score improvements of approximately 3.5 and 5.2, respectively, over GPT4. Additionally, PMC demonstrates superior performance compared to LES across both baseline systems. 
This is consistent with the results shown in Figure~\ref{fig:rate-valid} and Figure~\ref{fig:rate-faithful}. 
Similarly, LES and PMC exhibit improvements under the five-shot learning setting, indicating that the proposed methods remain effective even with a limited number of in-context demonstrations.

Additionally, it can be observed that PMC consistently achieves improvements over LLaMA3-70B-IT and Qwen2.5-72B-IT under the supervised fine-tuning setting. Furthermore, PMC shows larger improvements on out-of-domain datasets compared to in-domain datasets, which aligns with the observations discussed in Section~\ref{sec:our-res}. 


\begin{table*}[t]
	\centering
	\small
        \caption{Performance under zero-shot, five-shot and finetuning setting. $\uparrow$ indicates an improvement compared to the baseline. The best results are marked in bold.}
	\begin{tabular}{l|c|ccccc|c}
		\toprule
		\textbf{Model}&\textbf{PTB} &\textbf{Dialogue} & \textbf{Forum} & \textbf{Law}   & \textbf{Literature} & \textbf{Review} & \textbf{Avg}\\
		\midrule
		\multicolumn{8}{l}{\textit{\textbf{Zero-shot Learning}}} \\
            Qwen2.5-72B-IT   & 57.99 &55.83  & 53.14 & 53.59 & 51.28 & 53.14 & 54.22 \\
            \rowcolor{softblue}
		Qwen2.5-72B-IT-LES   & 60.43 & 58.73 & 56.22 & 57.71 & 51.30 & 57.69 & 57.01 ($\uparrow$2.79)  \\
            \rowcolor{softblue}
            Qwen2.5-72B-IT-PMC   & 62.10 & 61.78 & 57.31 & 59.03 & 53.76 & 59.31 & 58.88 ($\uparrow$4.66)  \\
            GPT4  & 73.00 & 65.74 &  61.02 &  70.39 &  60.34  &  63.31 & 64.16 \\
		\rowcolor{softblue}
		GPT4-LES  & 75.12 & 68.35 &  63.71 &  72.57 &  61.49  &  65.14 & 67.73 ($\uparrow$3.57) \\
		\rowcolor{softblue}
		GPT4-PMC  & \textbf{76.87} & \textbf{69.01} &  \textbf{64.78} &  \textbf{74.41} & \textbf{63.82}  &  \textbf{67.05} & \textbf{69.32} ($\uparrow$5.16) \\
		\midrule
	\multicolumn{8}{l}{\textit{\textbf{Five-shot Learning}}} \\
		Qwen2.5-72B-IT   & 62.13 & 60.23 &  56.72 & 60.38 & 52.49 & 58.43 & 58.39 \\
        \rowcolor{softblue}
        Qwen2.5-72B-IT-LES & 65.82 & 62.78 &  58.67 & 62.03 & 54.01 & 60.24 & 60.59 ($\uparrow$2.20)  \\
        \rowcolor{softblue}
        Qwen2.5-72B-IT-PMC   & 67.03 & 64.34 & 60.31 & 63.78 & 55.92 & 62.05 & 62.24 ($\uparrow$3.85) \\
        GPT4   & {73.28} & 66.61 &  62.01 & 72.43 & 58.29 &  66.41 & 65.15 \\
        \rowcolor{softblue}
        GPT4-LES  & {76.31} & 69.13 &  64.73 & 74.37 & 61.14 &  69.04 & 69.12 ($\uparrow$3.97)  \\
        \rowcolor{softblue}
        GPT4-PMC   & \textbf{78.57} & \textbf{70.23} &  \textbf{66.13} & \textbf{76.01} & \textbf{62.12} &  \textbf{70.87} & \textbf{70.66} ($\uparrow$5.51)  \\
        \midrule
        \multicolumn{8}{l}{\textit{\textbf{Supervised Fine-tuning}}} \\
        SEPar$\texttt{[graph]}$ & 95.72 & \textbf{86.30}  & \textbf{87.04} &  \textbf{92.06}  &  \textbf{86.26} & \textbf{84.34} &\textbf{88.62}  \\
        SAPar$\texttt{[graph]}$ & \textbf{96.40} & 86.01  & 86.17 &  91.71  &  85.27 & 83.41 & 88.16 \\
        LLaMA3-70B-IT & 95.88 &  {84.21} & 82.48  &   85.54    &   79.63  & 80.90 &  84.77 \\
        Qwen2.5-72B-IT  &{96.04}   &{84.20}   &{82.64}   &{85.73}      &{79.76}    &{81.41} &84.96 \\
        \rowcolor{softblue}
        LLaMA3-70B-IT-PMC & 96.07 & 85.42 & {84.53}  &  88.06    & 83.13  & 83.78 & 86.83 ($\uparrow$2.06) \\
        \rowcolor{softblue}
        Qwen2.5-72B-IT-PMC  &{96.28} &{85.78}  &{84.48}   &{88.41}   &{83.32}    &{84.12} &{87.07} ($\uparrow$2.11) \\
	\bottomrule
	\end{tabular}
	\label{tab:new-results-zs}
\end{table*}

\begin{figure}[!t]
    \centering
    \includegraphics[width=0.49\textwidth]{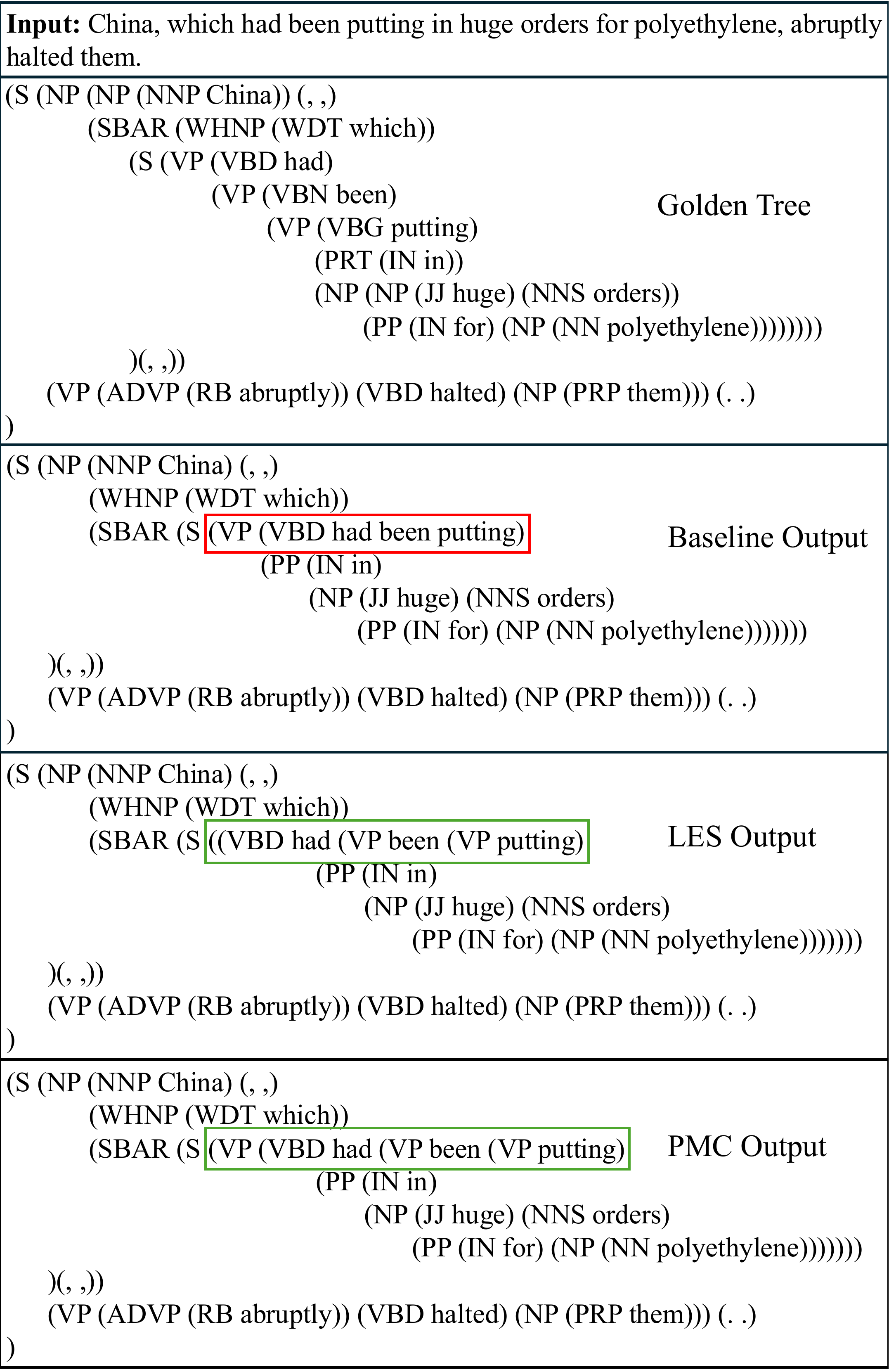}
    \caption{A sentence ``China, which had been putting in huge orders for polyethylene, abruptly
halted them.'' with its gold-standard constituent trees followed by constituent trees generated by three models.}
    \label{fig:case}
    \vspace{-0.3em}
\end{figure}

\begin{figure}[!t]
    \centering
    \includegraphics[width=0.49\textwidth]{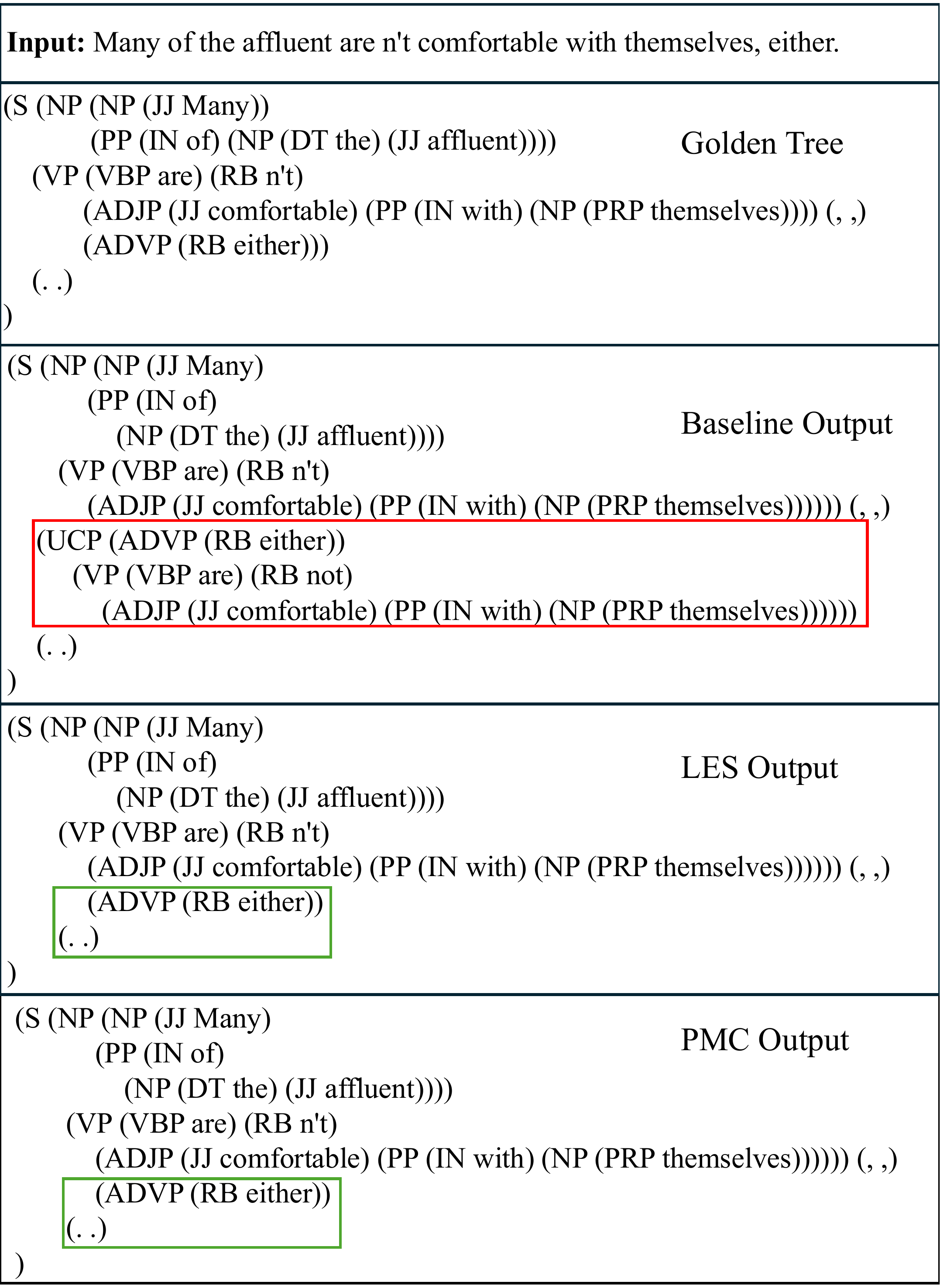}
    \caption{A sentence ``Many of the affluent are n't comfortable with themselves, either.'' with its gold-standard constituent trees followed by constituent trees generated by three models.}
    \label{fig:case2}
    \vspace{-0.3em}
\end{figure}

{\subsection{Comparison with the approach of Tian et al., (2024).}}
\label{sec:cmp-tian}
{We additionally compare our method with the three-stage LLM instruction framework proposed by Tian et al. (2024)~\cite{tian-etal-2024-large}. Notably, their approach utilizes the standard \textit{evalb} evaluation script, which excludes invalid parse trees from scoring. To ensure fairness, we report both the valid score measured by the standard evalb script and the overall score using our script, which accounts for these invalid constituents.}

\begin{table}[!t]
\centering
\small
{
\caption{Comparison with the approach of Tian et al., (2024) on PTB under a five-shot learning paradigm. $\dagger$ indicates our reproduced results.}}
\begin{tabular}{l|cc}
\toprule
 \textbf{Model} & \textbf{F1 (Valid)} & \textbf{F1 (Overall)} \\ 
\midrule
GPT4-3stage   &79.29 &- \\
GPT4-3stage$^{\dagger}$ &79.96 & 75.86 \\
\midrule
GPT4-LES &80.23 &76.31 \\
GPT4-PMC  &81.54 &78.57\\
GPT4-3stage-PMC &81.87 &79.52 \\
\bottomrule
\end{tabular}
\label{table:cmp-with-tian}
\vspace{-1.0em}
\end{table}

{Table~\ref{table:cmp-with-tian} compares the performance of the proposed methods and the GPT4-3stage approach\footnote{We also tried to reproduce the GPT4-3stage-COT method using the prompt template provided in the original paper. However, contrary to the reported results, our implementation did not yield better performance than the standard GPT4-3stage approach.} of Tian et al., (2024) on the PTB dataset under a five-shot learning paradigm. It can be observed that the proposed approaches obtain better results than GPT4-3stage, which verifies the effectiveness of the proposed method. 
Moreover, while GPT4-3stage achieves improvements over vanilla LLMs, we observe that it still encounters similar limitations, such as producing about 15\% structural invalid trees. In contrast, the proposed method can effectively reduce the invalid tree rate (about 8\%). This confirms that the contributions of the proposed methods are orthogonal to the work of Tian et al., (2024).}

{
To further verify the effectiveness of the proposed method, we integrate the approach of Tian et al. (2024) with our proposed method by employing PMC as a post-processing module for GPT4-3stage. As shown in Table~\ref{table:cmp-with-tian}, this combined approach yields further improvements over both individual methods, suggesting that the contributions of the two methods are complementary and orthogonal to each other.}

\subsection{Case Study}

To further understand the effectiveness of the proposed method, we conduct a case study on the Qwen2.5-72B-IT model under the five-shot setting.
As illustrated in Figure~\ref{fig:case}, the baseline output includes an invalid constituent, "\textit{(VBD had been putting)}," where three words are incorrectly grouped into a single constituent.
LES approach rectified this issue by separating "\textit{had been putting}" into three distinct constituents. 
Similarly, PMC also corrects the invalid constituent. Moreover, the constituent tree generated by PMC more closely resembles the golden tree compared to the one produced by LES.
These results confirm the effectiveness of our methods in mitigating the generation of invalid constituents. 

Figure~\ref{fig:case2} presents another case where the baseline output contains a nested constituent, ``\textit{(VP (VBP are) (RB not) (ADJP (JJ comfortable) (PP (IN with) (NP (PRP themselves))))))}'', which is unfaithful to the input sentence (i.e., the baseline repeatly outputs the constituent tree of the phrase ``are not comfortable with themselves''). 
It can be observed that both LES and PMC successfully mitigate this issue compared to the baseline. 
These results confirm that the proposed methods can effectively reduce the generation of unfaithful trees.

Due to space constraints, we present only two cases here, and additional cases will be included in the final version for a comprehensive analysis.

{\section{Discussion}}
{While the proposed methods demonstrate measurable improvements over baseline approaches, they are subject to two primary limitations: 1) Both LES and PMC frameworks incur substantially higher computational overhead compared to conventional LLMs. Notably, the PMC approach necessitates multi-agent interaction, which imposes significant additional inference costs; 2) The proposed methods primarily address structural invalidity and factual inaccuracies in generated trees, while exhibiting limited capability in identifying semantic-level inconsistencies (e.g., erroneous constituent labeling) and domain-specific anomalies. To enhance practical applicability, future research directions should focus on two key aspects: optimizing the computational efficiency of the error-processing pipeline and expanding the error taxonomy through automated detection methods. {In addition, future studies may explore integrating LLM predictions with graph-based parsers (e.g., SEPar) via re-ranking mechanisms to potentially enhance structural accuracy and length robustness.} Furthermore, given the inherent flexibility of the proposed frameworks, promising avenues for subsequent investigation include their adaptation to multilingual contexts and related computational linguistics tasks such as dependency parsing.}
\vspace{1em}
\section{Conclusion}
We investigated the usage of large language models (LLMs) for constituency parsing.
Through comprehensive evaluations involving six LLMs across six datasets under three different settings, 
we observed that 
1) LLMs obtain promising results on zero/few-shot constituency parsing;
2) In supervised learning scenarios, LLMs still lag behind graph-based parsers, likely due to their sequential nature and insufficient pre-training on structured outputs;
3) LLMs are prone to generating invalid and unfaithful constituent trees, which significantly limits their overall performance.
Motivated by these observations, we explored two approaches to mitigate the generation of invalid and unfaithful trees by LLMs.
Experimental results demonstrate that the proposed method can effectively reduce the occurrence of invalid and unfaithful trees, thereby improving overall parsing performance.
In the future, we would like to extend the proposed methods to enhance the out-of-domain generalization ability of LLMs. 

\ifCLASSOPTIONcaptionsoff
\newpage
\fi


\bibliographystyle{IEEEtran}
\bibliography{ref-sim}

\vfill
\end{document}